\documentclass[10pt,twocolumn,letterpaper]{article}

\usepackage{cvpr}              

\usepackage{multirow}

\usepackage{pifont}
\usepackage{graphicx}  
\usepackage{listings}
\usepackage{subcaption}   
\usepackage{cuted}
\usepackage{float}

\usepackage{xcolor} 
\usepackage{caption}    
\usepackage{soul}
\usepackage{xcolor}
\usepackage{pifont}
\usepackage{tikz}
\usepackage[utf8]{inputenc}
\usepackage{tcolorbox}
\tcbuselibrary{listings,breakable}
\usepackage{xcolor}
\usepackage{pifont}
\usepackage{geometry}
\usepackage{amsmath} 
\usepackage{amssymb} 
\usepackage{minted}
\usepackage{mdframed} 
\setminted{
  fontsize=\scriptsize,  
  baselinestretch=1.1,  
  breaklines=true,  
  framesep=2mm,  
  tabsize=2,  
  bgcolor=gray!10,  
}

\definecolor{customgreen}{rgb}{0.35, 0.65, 0.35}
\definecolor{customred}{rgb}{0.7, 0.35, 0.35}
\definecolor{customblue}{rgb}{0.3, 0.6, 0.8}
\definecolor{customblue1}{rgb}{0.3, 0.6, 0.8}
\definecolor{customblack}{rgb}{0.2, 0.2, 0.2}

\definecolor{darkred}{rgb}{0.6, 0, 0}
\definecolor{darkgreen}{rgb}{0, 0.5, 0}
\definecolor{darkblue}{rgb}{0, 0, 0.55}
\definecolor{cvprblue}{rgb}{0.21,0.49,0.74}

\newtcolorbox{promptbox}[1][]{
    colback=gray!10!white,
    colbacktitle=white,
    coltitle=black,
    colframe=black!75!black,
    boxrule=0.7pt,
    halign title=center,
    title=\textbf{#1}
}

\usepackage[pagebackref,breaklinks,colorlinks,allcolors=cvprblue]{hyperref}

\def\paperID{21042} 
\def\confName{CVPR}
\def\confYear{2026}

\title{
  VRAG-DFD: Verifiable Retrieval-Augmentation for MLLM-based Deepfake Detection
}

\renewcommand{\fnsymbol}[1]{%
  \ifcase#1\relax
    † 
  \or
    * 
  \else
    \fnsymbol{#1} 
  \fi
}

\author{
    Hui Han$^{1,2}$\thanks{Equal Contribution.}\footnotemark[1] \quad 
    Shunli Wang$^{2}$\footnotemark[1] \quad 
    Yandan Zhao$^2$ \\[2pt] 
    Taiping Yao$^2$ \quad 
    Shouhong Ding$^2$\thanks{Corresponding author (email: ericshding@tencent.com).} \\[5pt] 
    $^1$Shanghai Jiao Tong University \qquad
    $^2$Tencent Youtu Lab
    \vspace{.5em} 
    \\
    \textcolor{black}{\url{https://github.com/abigcatcat/VRAG-DFD.git}}
}

\begin{document}
\maketitle
\begin{abstract}

In Deepfake Detection (DFD) tasks, researchers proposed two types of MLLM-based methods: complementary combination with small DFD detectors, or static forgery knowledge injection.
The lack of professional forgery knowledge hinders the performance of these DFD-MLLMs.
To solve this, we deeply considered two insightful issues: How to provide high-quality associated forgery knowledge for MLLMs? AND How to endow MLLMs with critical reasoning abilities given noisy reference information?
Notably, we attempted to address above two questions with preliminary answers by leveraging the combination of Retrieval-Augmented Generation (RAG) and Reinforcement Learning (RL).
Through RAG and RL techniques, we propose the \texttt{\textbf{VRAG-DFD}} framework with accurate dynamic forgery knowledge retrieval and powerful critical reasoning capabilities.
Specifically, in terms of data, we constructed two datasets with RAG: Forensic Knowledge Database (FKD) for DFD knowledge annotation, and Forensic Chain-of-Thought Dataset (F-CoT), for critical CoT construction.
In terms of model training, we adopt a three-stage training method (Alignment$\to$SFT$\to$GRPO) to gradually cultivate the critical reasoning ability of the MLLM.
In terms of performance, \texttt{\textbf{VRAG-DFD}} achieved SOTA and competitive performance on DFD generalization testing. 

\end{abstract}    

\section{Introduction}




With the rapid advancement of Artificial Intelligence Generated Content (AIGC) technology, Deepfake techniques \cite{face2face, neuraltextures, faceshifter, diffswap} (\textit{e.g.}, face swapping and facial reenactment) and image generation methods \cite{GAN, LDM, DDPM} have already acquired the ability to produce high-fidelity forged images.
While such technologies demonstrate promising applications in entertainment media, film production, and other fields, they also pose severe security risks, including privacy infringement, financial fraud, and disinformation dissemination. 
As a core defensive measure, Deepfake Detection (DFD) technology has grown increasingly critical for ensuring the credibility of visual content in the era of AIGC-driven forgery.

\begin{figure}[tb]  
  \centering  
  \includegraphics[width=0.5\textwidth]{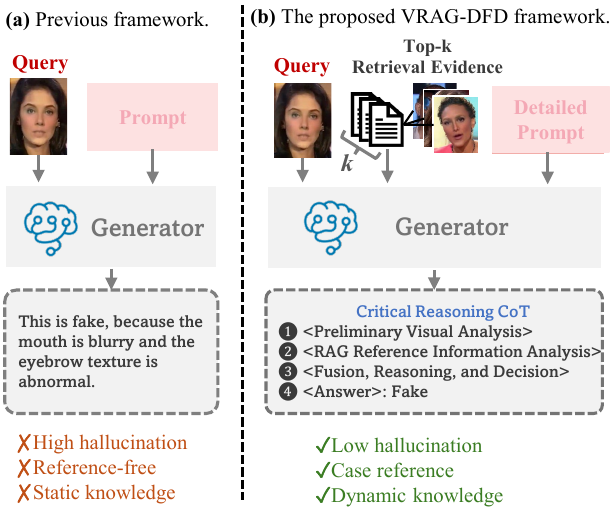}  
  \vspace{-15pt}
  \caption{Comparison between the previous method and the proposed \texttt{\textbf{VRAG-DFD}} framework. Previous methods predicted based on a single image and prompt with severe hallucinations. The proposed  \texttt{\textbf{VRAG-DFD}} effectively reduce hallucinations by constructing highly relevant references through RAG and making decisions through critical reasoning. (Faces are selected from publicly available FF++ \cite{FF++} dataset.)}
  \label{fig1} 
\end{figure}



Traditional deepfake detectors \cite{li2020face, SBI, fu2025exploring, zhu2021face, cao2022end, khalid2020oc, Seeable, effort} rely on fine-tuning visual backbone networks (\textit{e.g.}, CNNs or ViTs) and enhance generalizability through delicate network designs. 
These face forgery detectors typically aim to mine discriminative fake patterns (\textit{e.g.}, inconsistent facial textures, abnormal lighting) and thereby distinguish fake images \cite{effort}.
As simple binary classifiers, these detectors can only output probabilistic results and cannot generate interpretable text, thus failing to meet users’ interactive needs, such as the demand for detailed analysis on \textit{Why an image is classified as fake?} \cite{X2_dfd}.

The emergence of Multimodal Large Models (MLLMs) \cite{qwen25} has offered new prospects for addressing the aforementioned challenges.
Some existing research \cite{X2_dfd, TruthLens, VLF_FFD, M2F2_Det, KFD} has already integrated MLLMs into Deepfake detection tasks, enabling these detectors to generate textual explanations and thus enhancing model interpretability.
Some models \cite{X2_dfd, TruthLens, VLF_FFD} adopt a hybrid framework combining MLLM and small DFD detectors, enabling a clear division of labor between \textit{High-level} semantic analysis and \textit{Low-level} artifacts extraction.
Other models \cite{M2F2_Det, KFD} injected DFD expertise through sparse static external knowledge such as isolated keyword or attribute lists. 
We argue that the combination strategy and static knowledge injection approach cannot enable MLLM to have true understanding, criticism, and thinking abilities in DFD task, which means that the \textit{\textbf{Knowledge Gap}} problem still exists.
To solve this, we need to address two unavoidable issues: 

\textbf{\textit{Q1: How to provide high-quality detailed associated DFD knowledge for MLLM ?}}

\textbf{\textit{Q2: How to endow MLLM with both critical reasoning abilities and the capacity to recognize the associated knowledge?}}




For \textit{\textbf{Q1}}, we adopt Retrieval-Augmented Generation (RAG) to solve the detailed associated DFD knowledge issue.
The existing annotation methods \cite{M2F2_Det, KFD} usually start with designing a detailed annotation prompt, and then use closed or open source LLMs to annotate individual fake images.
Assuming this process is denoted as $P (T|M,I)$, where $M$ denotes the annotator, $I$ denotes the single target image, $T$ denotes the annotation text.
We believe that a feasible solution to achieve better deepfake knowledge is to provide a series of $K$ highly relevant cases for auxiliary annotation.
The annotation process becomes $P(T|M,I,\{I_k\}_{1}^{K}, \{T_k\}_{1}^{K})$.
Specifically, we constructed Forensic Knowledge Database (FKD) and Forensic Chain-of-Thought (F-CoT) Datsets to obtain more reliable associated DFD annotations.


For \textit{\textbf{Q2}}, we propose \texttt{\textbf{VRAG-DFD}}, an interpretable Deepfake detection framework based on RAG and critical thinking training, as shown in Figure \ref{fig1}. 
Specifically, based on the above reliable annotations, we designed a three-stage training strategy to gradually inject critical reasoning abilities into MLLMs.
(1) Firstly, we equip the MLLM with basic visual perception capabilities through alignment learning on large-scale dataset.
(2) Secondly, we adopted a high-quality forensic Chain-of-Thought (F-CoT) dataset to fine tune the model's instructions, with the goal of equipping the model with critical thinking ability. The model is able to cross validate \textbf{\textit{External Retrieval Evidence}} with its own \textbf{\textit{Internal Visual Analysis}}. 
(3) Finally, this complex evidence balancing and reasoning ability is further refined through reinforcement learning \cite{deepseek}. 
Our main contributions are threefold.

\begin{itemize}
\item We proposed high-quality annotation method with retrieval strategy and construct a detailed Forensic Knowledge Database (FKD) and a innovative critical thinking instruction datasets (F-CoT).
\item By combining RAG and RL techniques, we designed a MLLM-based critical thinking framework named \texttt{\textbf{VRAG-DFD}}. We have explored an effective training strategy to cultivate critical reasoning skills: Alignment$\to$SFT$\to$GRPO.

\item The proposed framework achieves SOTA \& competitive generalization performance on multiple deepfake detection benchmarks with excellent interpretation fidelity.
\end{itemize}

\section{Related work}
\subsection{Traditional Deepfake Detection}

Traditional Deepfake detectors typically adopt CNNs and ViTs as basic frameworks, and employ various data processing and transformation methods for forgery detection, such as frequency transformation \cite{liu2021spatial, luo2021generalizing, li2021frequency}, blending operations \cite{li2020face, SBI, fu2025exploring}, reconstruction methods \cite{zhu2021face, cao2022end}, and learning real distribution with anomaly detection \cite{khalid2020oc, Seeable}.
Although these strategies have achieved some performance improvements, there are still serious generalization issues.
Recently, some methods achieved high generalization by retaining high-quality visual-language pre-trained knowledge.
Effort \cite{effort} employes singular value decomposition to decompose the original CLIP features into semantic and artifacts space, thus preserving the pre-trained knowledge while learning fake patterns.
Despite their intricate structural design, the essence of these detectors is still a "black box" for direct binary classification, and the output results lack interpretability.
Therefore, combining dedicated Deepfake detectors and MLLM is a very intuitive way to improve the interpretability.

\subsection{MLLMs for Deepfake Detection}
In the Deepfake detection task, the introduction of multimodal large models (MLLMs) provides the possibility for determining generalization and interpretability \cite{X2_dfd, TruthLens, VLF_FFD}. These models are capable of analyzing and inferring image content through language, thereby identifying forgery content.
Although MLLM excels in semantic understanding, it requires professional low-level information guidance in such heavy trace DFD detection tasks.
To achieve this goal, VLF-FFD \cite{VLF_FFD}, $\chi^2$-DFD \cite{X2_dfd}, TruthLens \cite{TruthLens} adopted small deepfake models for information complementarity. They compensate the main MLLM by supplementing with external detectors. But this approach is redundant in architecture, and knowledge remains static after training.
On the other hand, KFD \cite{KFD} and M2F2-Det \cite{M2F2_Det} inject a general forgery knowledge into LLM through prompt learning methods, but the information content injected by their knowledge is also isolated and low-quality.
Unlike these methods, this research abandons the static generic forgery trace list and instead uses the RAG strategy for dynamic retrieval and annotation. This method provides more flexible, precise, and targeted external knowledge.


\begin{figure*}[tb]  
  \centering  
  \includegraphics[width=1.0\textwidth]{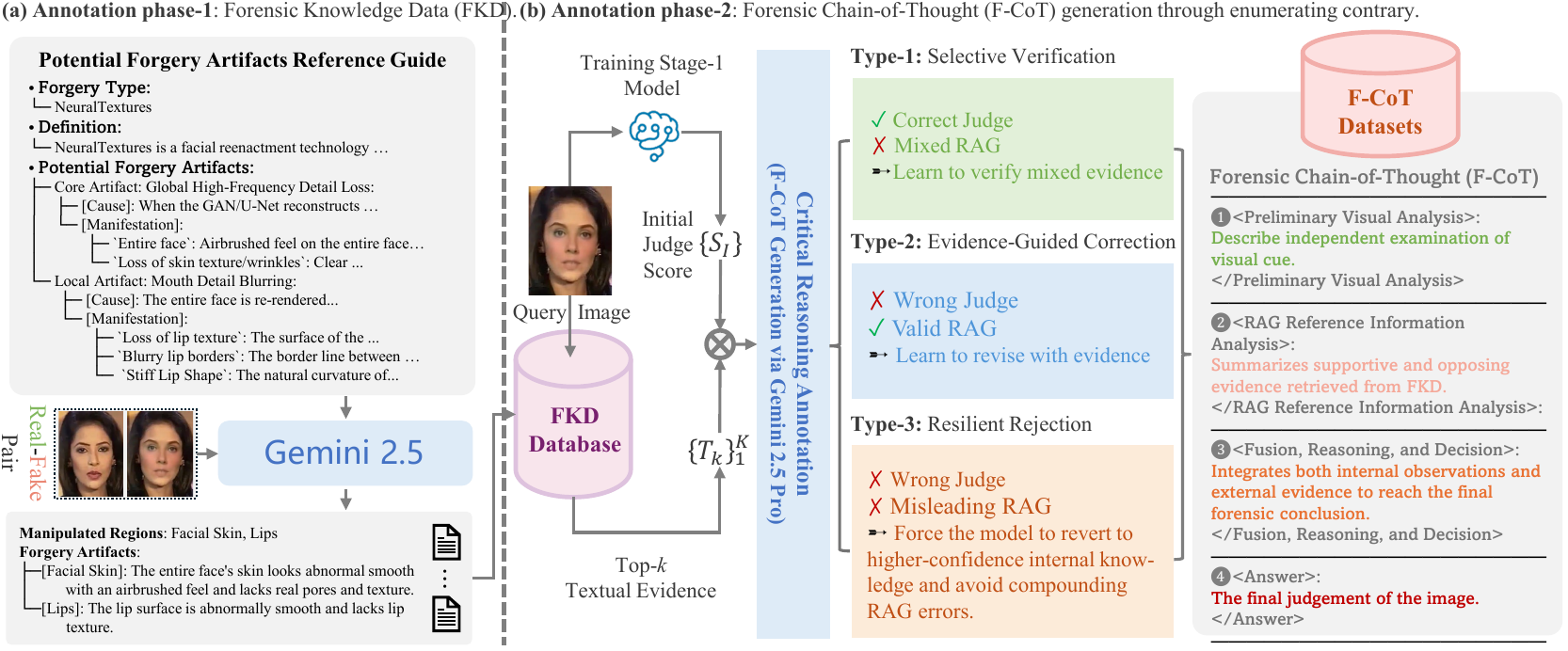}  
  \caption{The anotation phases of the Forensic Knowledge Database (FKD) and Forensic Chain-of-Thought (F-CoT) Dataset. (Faces are selected from FF++ \cite{FF++} dataset.)}
  \label{fig2} 
\end{figure*}

\subsection{Retrieval-Augmented Generation}
Retrieval-Augmented Generation (RAG) is a pivotal technology designed to facilitate external knowledge expansion for existing Large Language Models (LLMs) \cite{He2022RethinkingWR}. 
By leveraging relevant information retrieved from external knowledge repositories (\textit{e.g.}, academic databases, structured knowledge graphs), RAG can significantly mitigate the occurrence of hallucinations during the model's generation process \cite{Raunak2021TheCC, ji2023survey}. This capability enables LLMs to produce more accurate, factually consistent, and reliable responses.
At present, RAG has been widely applied in fields such as code generation \cite{zhou2022docprompting}, Visual Question Answering (VQA) \cite{{dahl2024large,pu2024autorepo}}, and video understanding \cite{arefeen2024irag,tan2025rag}. 
This paper firstly introduce RAG technology into Deepfake detection tasks.
Experimental results confirm that the integration of RAG significantly reduces the hallucination during the annotation phase, ultimately contributing to the construction of a more accurate and trustworthy DFD detector.


\section{Method}
\subsection{Overview}



Our VRAG-DFD approach consists of three key components. First, we build two datasets: a Forensic Knowledge Dataset (FKD) with fine-grained artifact annotations and a Critical Reasoning Dataset with structured F-CoT annotations. Second, the framework integrates a Dynamic Forensic Retriever and a Critical Reasoning MLLM to retrieve relevant evidence and perform multi-step forensic reasoning. Finally, a three-stage training pipeline foundational visual alignment training, F-CoT supervised fine-tuning, and RL-GRPO autonomous reasoning reinforcement progressively equips the model with forensic perception, critical reasoning, and autonomous decision-making capabilities.


\subsection{Datasets}
The foundation of our method lies in two hierarchically complementary datasets: an external knowledge base for RAG retrieval, and an instruction dataset for training critical reasoning.
Two annotation phases are demonstrate in Figure \ref{fig2}.

\subsubsection{Forensic Knowledge Database}

The forensic reasoning capability of \textbf{VRAG-DFD} heavily depends on the quality of evidence stored in its external knowledge base. To support reliable retrieval, we construct a high-quality, fine-grained \textbf{Forensic Knowledge Database (FKD)} with detailed expert-level annotations.

We use the FaceForensics++ (FF++) \cite{FF++} dataset as the source data for building FKD. The database contains approximately 18k entries, with a nearly balanced ratio of forged and real samples (1:1). Each entry is formatted as \texttt{[forged region]: "description of forgery artifacts..."}, enabling structured knowledge retrieval and reasoning.

To ensure annotation precision and forensic reliability, we employ a \textbf{guide-driven contrastive annotation} strategy:

\begin{itemize}
    \item \textbf{Forgery Guides:} Based on a comprehensive literature review, we design detailed \emph{Forgery Guides} for the four manipulation techniques in FF++ \cite{FF++} (DeepFakes \cite{DeepFakes}, Face2Face \cite{face2face}, FaceSwap \cite{FaceSwap}, and NeuralTextures \cite{neuraltextures}). Each guide defines the characteristic manipulation patterns of its corresponding method and describes its typical visual artifact, such as boundary blurring, texture inconsistency, and illumination anomalies using precise forensic terminology.

    \item \textbf{Contrastive Annotation:} We employ \emph{Gemini 2.5 Pro} as the annotation assistant. During annotation, both forged and pristine frames are presented simultaneously, allowing direct comparison. This contrastive setup facilitates accurate localization of manipulated regions and ensures that artifact descriptions strictly follow the \emph{Forgery Guides}.
\end{itemize}

\subsubsection{Critical Reasoning Dataset}

To enable the MLLM to \emph{critically} utilize RAG-retrieved evidence, we construct an instruction-tuning dataset specifically designed to teach \textbf{forensic critical reasoning}.

\noindent\textbf{Data Source.} We sample 9,000 image instances from FF++ \cite{FF++} as the visual inputs. For each image, we additionally retrieve $k$ textual evidence entries from the Forensic Knowledge Database (FKD), which together form the input pairs used for this fine-tuning stage.

\noindent\textbf{Forensic Chain-of-Thought (F-CoT) Generation.} For each sample, we first obtain an \emph{initial judgment} from the Stage-1 trained model and simultaneously retrieve the top-$k$ external evidence entries from the Forensic Knowledge Database (FKD). Based on the correctness of the model’s initial judgment and the validity of the retrieved RAG evidence, we categorize all samples into three types. Each type is then annotated with a \textbf{Forensic Chain-of-Thought (F-CoT)} generated by \emph{Gemini 2.5 Pro}, serving as the gold-standard reasoning supervision:

\begin{itemize}
    \item \textbf{Type 1 (Concurrence / Cross-Verification):} [Correct initial judgment, mixed RAG evidence].  
    F-CoT annotations in this category teach the model how to perform cross-verification among mixed-quality evidence. The model learns to identify and adopt evidence that supports its correct reasoning while rejecting misleading counter-evidence, ultimately reinforcing its initial conclusion.

    \item \textbf{Type 2 (RAG-Corrective / Evidence-Guided Correction):} [Incorrect initial judgment, valid RAG evidence].  
    These samples train the model to recognize when its initial judgment is incorrect and to leverage valid RAG evidence for correction. The model is guided to analyze specific artifact-level cues provided by RAG (\textit{e.g.}, \texttt{[mouth region]: abnormal blending}) to verify them against the image and revise its final decision accordingly.

    \item \textbf{Type 3 (Adversarial / Resilient Rejection):} [Incorrect initial judgment, misleading RAG evidence].  
    This type provides the most challenging and adversarial cases. The F-CoT annotations guide the model to exercise critical reasoning: first identifying that the retrieved evidence is invalid or misleading (\textit{e.g.}, the retrieved "similar" samples are only semantically similar), then \textbf{overriding} such evidence and reverting to high-confidence internal forensic priors (\textit{e.g.}, “the philtrum structure is intact and biologically consistent, hence the face is authentic”) to arrive at a correct final judgment.
\end{itemize}

\noindent\textbf{F-CoT Response Structure.} 
All F-CoT annotations adhere to a unified and structured response format that explicitly separates visual observation, evidence analysis, and final decision-making. This structure guides the model to produce interpretable and logically consistent forensic reasoning:













\begin{itemize}
  \item \textbf{Preliminary Visual Analysis} describes the model’s independent examination of visual cues (\textit{e.g.}, texture consistency, artifacts, anomalies).
  \item \textbf{RAG Reference Information Analysis} summarizes supportive and opposing evidence retrieved from FKD, along with critical verification of their reliability.
  \item \textbf{Fusion, Reasoning, and Decision} integrates both internal observations and external evidence to reach the final forensic conclusion.
  \item The \textbf{Ground\_Truth\_Label} denotes the verified authenticity of the sample (\textit{e.g.}, “Real” or “Fake”).
\end{itemize}

\subsection{VRAG-DFD Framework Architecture}
Our reasoning framework consists of two core components: a \textit{\textbf{Dynamic Forensic Retriever}} and a \textit{\textbf{Critical Reasoning MLLM}}. The former retrieves relevant forensic evidence from an external knowledge base, while the latter performs high-level reasoning and decision-making based on both the visual input and retrieved evidence.

\begin{figure*}[htb]  
  \centering  
  \includegraphics[width=1.0\textwidth]{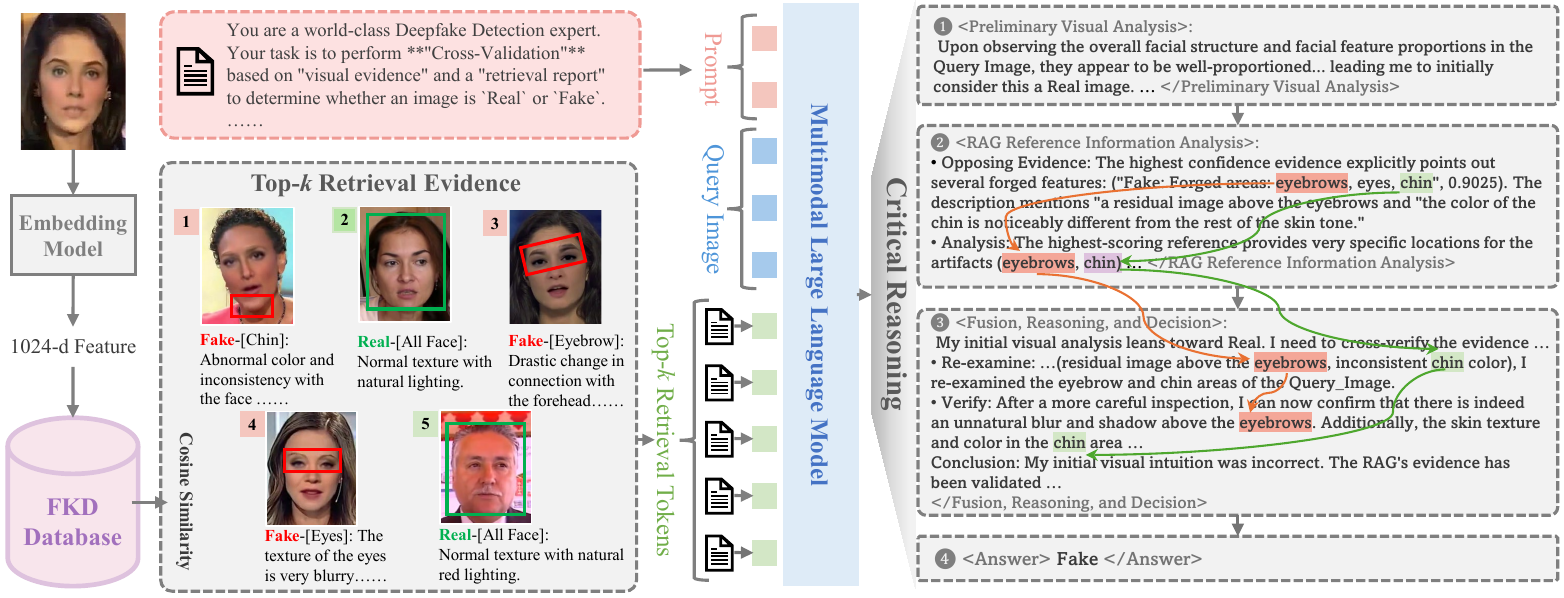}  
  \caption{Framework of the proposed VRAG-DFD. Firstly, the dynamic forensic retriever captures top-$k$ samples from the multimodal FKD database. Afterwards, MLLM conducts a critical reasoning process (\textit{e.g.}, \ding{172}-\ding{173}-\ding{174}) based on the text information retrieved by RAG , and ultimately outputs the final conclusion (\textit{e.g.}, \ding{175}). (Faces are selected from FF++ \cite{FF++} dataset.)}
  \label{fig3} 
\end{figure*}

\subsubsection{Dynamic Forensic Retriever}
The goal of this retriever is to dynamically retrieve cases from an external knowledge base Forensic Knowledge Database (FKD) that are highly similar to the query image.
Considering that general pre-trained models (\textit{e.g.}, CLIP-ViT \cite{clip}, ResNet \cite{resnet}) cannot effectively distinguish deepfake images, we adopted traditional DFD detector \cite{effort} as the embedding model to achieve this retrieval function.
The detailed retrieval process is shown in Figure \ref{fig3}.



\subsubsection{Critical Reasoning MLLM} 

We employ a pretrained multimodal large language model (MLLM) as the core reasoning module. During inference, the model takes as input a query image and prompts. More importantly, it also receives the Top-$k$ external evidence retrieved by the Forensic Retriever. Notably, the retrieved evidence is not the similar images themselves but the high-quality \emph{F-CoT textual annotations} associated with them. 

We deliberately choose to use textual evidence rather than raw image inputs for two reasons: 
(1) it prevents redundant visual noise and reduces computational overhead during multimodal reasoning, and 
(2) F-CoT annotations explicitly encode expert-level forensic interpretations (\textit{e.g.}, manipulation regions and their corresponding manipulation artifacts.), which are more informative and interpretable for reasoning than pixel-level similarity alone. 
The key idea of our approach is to leverage image similarity retrieval to dynamically obtain textual evidence that is visually relevant to the query case. 
The MLLM then critically compares and integrates these retrieved F-CoT evidences with the visual content of the query image to perform forensic reasoning and make the final authenticity judgment.
\begin{figure*}[htb]  
  \centering  
  \includegraphics[width=1.0\textwidth]{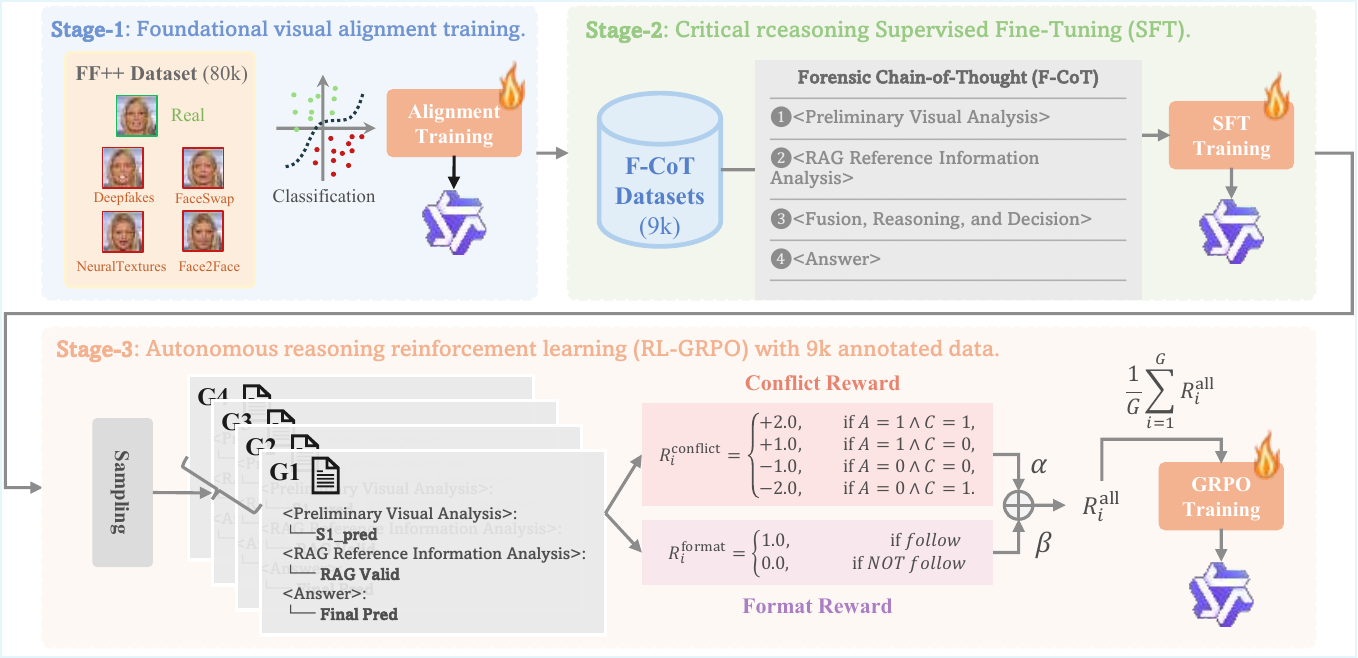}  
  \vspace{-15pt}
  \caption{Training stages of VRAG-DFD. Stage-1 aims to equip the MLLM with basic forgery artifacts recognition capabilities. Stage-2 aims to guide the model to predict in the form of F-CoT. Stage-3 adopts GRPO with detailed reward function designs to encourage the MLLM to have the capabilities to critical reasoning given top-$k$ RAG evidence. (Faces are selected from FF++ \cite{FF++} dataset.)}
  \label{fig4} 
\end{figure*}


\subsection{Three-Stage Training Pipeline}

To progressively unlock the model's forensic reasoning capabilities, we design the three-stage training pipeline illustrated in Figure \ref{fig3}.

\subsubsection{Stage-1: Foundational Visual Alignment}

\noindent\textbf{Goal.} 
This stage aims to establish the model's foundational visual perception abilities. Unlike Stage-2, which relies on fine-grained annotations, Stage-1 leverages large-scale, relatively simple labeled data for coarse-tuning, thereby laying a solid visual foundation for subsequent complex forensic reasoning tasks.

\noindent\textbf{Data.} 
We use all frames from 2,500 videos in FF++, totaling approximately 80k images. No RAG knowledge base (Sec.~3.2.1) or F-CoT annotations (Sec.~3.2.2) are employed at this stage.

\noindent\textbf{Task.} 
The model is fine-tuned on a standard Visual Question Answering (VQA) task with only the basic classification labels. Experiments demonstrate that large-scale VQA pre-training significantly enhances the model's visual representation capability, which is essential for Stage-2 when aligning visual features with fine-grained forensic artifacts.

\subsubsection{Stage-2: Critical Reasoning SFT}

\noindent\textbf{Goal.} 
Through imitation learning, this stage teaches the model to execute the three complex \emph{Forensic Chain-of-Thought (F-CoT)} reasoning patterns defined in Sec.~3.2.2.

\noindent\textbf{Data.} 
We use the \emph{Critical Reasoning Dataset} constructed in Sec.~3.2.2.

\noindent\textbf{Task.} 
This stage forms the core of training the model’s critical reasoning skills. Building on the visual capabilities from Stage-1, the model is supervised with gold-standard F-CoT annotations. The model is trained to emulate a structured forensic reasoning procedure. Specifically, the output is formatted as:

\begin{enumerate}
    \item \textbf{\texttt{<Preliminary Visual Analysis>}}: Independently analyze the query image based on Stage-1 visual capabilities and provide an initial judgment.
    \item \textbf{\texttt{<RAG Reference Information Analysis>}}: Analyze the Top-$k$ textual evidence retrieved from the Forensic Retriever (Sec.~3.3.1), clearly distinguishing between \emph{supportive} and \emph{opposing} evidence.
    \item \textbf{\texttt{<Fusion, Reasoning, and Decision>}}: Perform the core critical reasoning step: cross-verify the validity of the RAG evidence against the image, weigh the initial judgment and external evidence, and produce an integrated final reasoning process.
    \item \textbf{\texttt{<Answer>}}: Output the final label: Real or Fake.
\end{enumerate}

By imitating the three data types defined in Sec.~3.2.2, the model learns to handle conflicting evidence and to master the “forensic-verification” F-CoT reasoning structure.

\subsubsection{Stage-3: Autonomous Reasoning Reinforcement (RL-GRPO)}

\noindent\textbf{Goal.} 
The goal of Stage-3 is to elevate the imitation capabilities learned in Stage-2 to true \textit{\textbf{Autonomous Decision-Making}} and \textit{\textbf{Critical Thinking}}. Specifically, we aim to encourage the model to actively explore reasoning strategies when encountering conflicting evidence, rather than merely reproducing the SFT demonstration paths.

\noindent\textbf{Data.} 
To mitigate potential overfitting from SFT (\textit{e.g.}, Advantage $\approx 0$), Stage-3 is conducted on data unseen during Stage-1 and Stage-2. This ensures sufficient optimization space for exploring new data distributions. We use previously unobserved frames from FF++ \cite{FF++} as the training set.

\noindent\textbf{Process-Aware Reward Function.} 
The Process-Aware Reward Function as the sum of two components: Format Reward  and Conflict Reward.
Format Reward mandates the model to follow the \textbf{F-CoT} structure.
Conflict Reward dynamically evaluates the reasoning paths generated by the policy model  relying on two key signals: 

1. The model’s preliminary visual analysis, which reveals its inherent preference and initial prediction ($s1\_pred$).

2. The validity of the retrieved RAG evidence (\textit{e.g.}, assessed via majority voting or annotation consistency), which reflects the reliability of external knowledge.

By comparing $s1\_pred$ with the RAG evidence, we can detect conflicts: if the model correctly resolves conflicts, it demonstrates successful integration of internal visual knowledge and external evidence, exhibiting true critical reasoning; if the model fails, producing an incorrect prediction, it indicates potential over-reliance or over-rejection of RAG evidence.

Specifically, we define $A \in \{0, 1\}$ as an indicator of whether the model’s final answer is correct ($A=1$ denotes correctness). We further define $C \in \{0, 1\}$ to represent whether there exists a conflict between the model’s preliminary visual judgment ($s1\_\text{pred}$) and the evidence retrieved via RAG ($rag\_\text{correct}$), where $C=1$ indicates a conflict. 
Our conflict reward function $\mathcal{R}_{\text{conflict}}$ is formulated as follows:
\begin{equation}
\mathcal{R}_{\text{conflict}} =
\begin{cases}
+2.0, & \text{if } A=1 \wedge C=1 \\
+1.0, & \text{if } A=1 \wedge C=0 \\
-1.0, & \text{if } A=0 \wedge C=0 \\
-2.0, & \text{if } A=0 \wedge C=1
\end{cases}
\end{equation}

This incentive structure creates a clear reinforcement loop:
\begin{itemize}
    \item \textbf{Highest reward (+2.0):} Assigned to \emph{critical correctness}, where the model correctly resolves a conflict ($C=1$) between internal perception and external evidence ($A=1$).
    \item \textbf{Lowest reward (-2.0):} Strongly penalizes \emph{critical errors}, where the model fails under conflicting evidence ($C=1$, $A=0$).
\end{itemize}

Through this design, the model is explicitly encouraged during the GRPO stage to explore and reinforce high-value critical reasoning behaviors, rather than merely repeating the \emph{simple correctness} patterns learned during the SFT stage.



\begin{table*}[htp]
  \centering
  \caption{Cross-dataset performance comparison with SOTA face forgery detection methods on CDF-v1 \cite{cdf}, CDF-v2 \cite{cdf}, DFDC \cite{dfdc}, FFIW \cite{ffiw}, and WDF \cite{wdf}. All methods are trained on FF++ \cite{FF++}. The best results are shown in \textbf{bold}, and the second results are \underline{underlined}. The Video-level AUC metric is adopted for comparison.}
  \vspace{-5pt}
  \footnotesize  
  \renewcommand\tabcolsep{5.0pt}  
  \renewcommand\arraystretch{1.0}  
  \begin{tabular}{ll|c|l| c c c c c}  
    \toprule  
    \textbf{Method} & \textbf{Venue} & \textbf{Explainable} & \textbf{Research Topic} & \textbf{CDF-v1} & \textbf{CDF-v2} & \textbf{DFDC} & \textbf{FFIW} & \textbf{WDF} \\
    \midrule  

    F$^3$-Net \cite{f3Net} & ECCV 2020      & \ding{55} & \footnotesize{Frequency + CNN} & 81.11 & 77.92 & 67.35 & 70.11 & 72.80 \\
    LTW \cite{LTW} & AAAI 2021              & \ding{55} & \footnotesize{Meta Training} & - & 77.14 & 69.00 & 76.63 & - \\
    SPSL \cite{spsl} & CVPR 2021            & \ding{55} & \footnotesize{Phase Spectrum} & - & 79.90 & 77.00 & 79.40 & 70.20 \\
    SRM \cite{SRM} & CVPR 2021              & \ding{55} & \footnotesize{Attention + FPN} & - & 84.00  & 69.50 & 80.60 & 72.20 \\
    PCL+I2G \cite{PCLI2G} & ICCV 2021       & \ding{55} & \footnotesize{Blending} & - & 90.03 & 67.52 & - & - \\
    SBI \cite{SBI} & CVPR 2022              & \ding{55} & \footnotesize{Blending} & 93.44 & 93.18 & 72.42 & 84.83 & - \\
    CORE \cite{CORE} & CVPR 2022            & \ding{55} & \footnotesize{Augmentation} & - & 80.90 & 72.10 & 71.00 & 72.40 \\ 
    DCL \cite{DCL} & AAAI 2022              & \ding{55} & \footnotesize{Contrastive Learning} & - & 82.30 & 76.71 & 71.14 & 71.14 \\
    SeeABLE \cite{Seeable} & ICCV 2023      & \ding{55} & \footnotesize{Contrastive Learning} & - & 87.30 & 75.90 & - & - \\  
    F$^2$Trans \cite{F2trans} & TIFS 2023   & \ding{55} & \footnotesize{Attention + Wavelet} & 86.29 & 89.87 & - & - & - \\  
    AltFreezing \cite{Altfreezing} & CVPR 2023     & \ding{55} & \footnotesize{Spatial + Temporal} & 88.48 & 89.50 & 64.75 & - & - \\ 
    AUNet \cite{AUNet} & CVPR 2023          & \ding{55} & \footnotesize{Facial AU} & - & 92.77 & 73.82 & 81.45 & - \\  
    UCF \cite{ucf} & ICCV 2023              & \ding{55} & \footnotesize{Disentanglement} & 86.08 & 83.73 & 75.11 & 69.70 & 77.42 \\ 
    LAA-Net \cite{LAA} & CVPR 2024          & \ding{55} & \footnotesize{Attention + FPN} & - & 95.40 & - & - & - \\  
    FreqBlender \cite{FreqBlender} & NeurIPS 2024   & \ding{55} & \footnotesize{Frequency + Blending} & - & 94.59 & 74.59 & 86.14 & - \\  
    UDD \cite{UDD} & AAAI 2025              & \ding{55} & \footnotesize{Token Shuffle + Mixing} & - & 93.10 & 81.20 & - & - \\ 
    LESB \cite{LESB} & WACVW 2025           & \ding{55} & \footnotesize{Blending} & - & 93.13 & 71.98 & 83.01 & - \\  
    Effort \cite{effort} & ICML 2025        & \ding{55} & \footnotesize{Fine-tune CLIP-ViT} & 96.05 & \underline{95.60} & \underline{84.30} & \underline{92.10} & \underline{84.80} \\ 
    \midrule  
    $\chi^2$-DFD \cite{X2_dfd} & NeurIPS 2025   & \ding{51} & \footnotesize{MLLM + Detector} & - & 95.50 & \textbf{85.30} & 86.70 & 86.40 \\  
    KFD \cite{KFD} & ICML 2025                  & \ding{51} & \footnotesize{MLLM + Static Knowledge} & \underline{97.62} & 94.71 & 79.12 & - & - \\  
    M2F2-Det \cite{M2F2_Det} & CVPR 2025        & \ding{51} & \footnotesize{MLLM + Static Knowledge}& - & 95.10 & - & 88.70 & - \\
    \midrule  
    \textbf{Ours} & CVPRF 2026 & \ding{51} & \footnotesize{MLLM +\textbf{ Dynamic RAG}} & \textbf{99.60} & \textbf{95.97} & 81.82 & \textbf{93.49} & \textbf{88.96} \\ 
    \bottomrule  
  \end{tabular}
  \label{tab:dfd_performance}  
\end{table*}

For each sample \(i\), we define the final reward as the sum of a conflict reward and a format reward:

\begin{equation}
R_i = \alpha \, r_i + \beta \, f_i.
\end{equation}
with \(\alpha\) and \(\beta\) denoting the weighting coefficients for the respective reward components.

For a batch of \(G\) samples, we take the mean reward as the batch-level supervision signal:
\begin{equation}
R = \frac{1}{G} \sum_{i=1}^{G} \left( \alpha \, r_i + \beta \, f_i \right).
\end{equation}

This design rewards successful conflict resolution and critical reasoning.


\section{Experiment}

\subsection{Basic Settings}

\noindent\textbf{Datasets.}
All training stages are conducted on FaceForensics++ (FF++) dataset \cite{FF++}. 
Following the split method with \cite{DF40}, we keep the same data partitioning settings in all three training stages. 
Specifically, Stage-1 (Alignment FT) utilize 2,500 videos from the FF++ training set. Stage-2 (Critical SFT) and Stage-3 (RL-GRPO) are trained on the remaining 1,000 videos. 
To rigorously evaluate the generalization capability of the proposed method, we adopt a strict cross-dataset evaluation setting: no images or videos from any of the evaluation datasets are used during any SFT stage. 
We report performance on five widely adopted Deepfake detection benchmarks: Celeb-DF-v1 (CDF-v1) \cite{cdf}, Celeb-DF-v2 (CDF-v2) \cite{cdf}, FFIW \cite{ffiw}, DFDC \cite{dfdc} and WildDeepfake (WDF) \cite{wdf}. 
These datasets cover diverse manipulation types, acquisition conditions, and levels of visual realism, enabling a comprehensive assessment of robustness and transferability.

\noindent\textbf{Baselines}. As shown in Table \ref{tab:dfd_performance}, we compare VRAG-DFD with two categories of state-of-the-art methods. 
Traditional SOTA detectors contain F$^3$-Net \cite{f3Net}, LTW \cite{LTW}, SPSL \cite{spsl}, SRM \cite{SRM}, PCL+I2G \cite{PCLI2G}, DCL \cite{DCL}, SBI \cite{SBI}, CORE \cite{CORE}, F$^2$Trans \cite{F2trans}, AUNet \cite{AUNet}, AltFreezing \cite{Altfreezing}, SeeABLE \cite{Seeable}, UCF \cite{ucf}, LAA-Net \cite{LAA}, FreqBlender \cite{FreqBlender}, UDD \cite{UDD}, LESB \cite{LESB} and Effort \cite{effort}. MLLM-based Detectors contain $\chi^2$-DFD \cite{X2_dfd}, KFD \cite{KFD} and M2F2-DET \cite{M2F2_Det}.

\noindent\textbf{Evaluation Metrics.} We use the Area Under the Receiver Operating Characteristic Curve (AUC) as the primary evaluation metric for classification and generalization performance. 
Following \cite{M2F2_Det}, we adopted GPT-4o as the automatic judge
to assess the quality of generated explanations. 
Detailed qualitative comparisons are provided in the supplementary materials.

\noindent\textbf{Implementation Details.} The proposed VRAG-DFD framework is based on the Qwen2.5-VL MLLM \cite{qwen25}. 
Low-Rank Adaptation (LoRA) \cite{lora} is adopted for all training stages with $r=128, \alpha=256$.
Training configurations of three stages are summarized in Table \ref{training_config}. 


\begin{table}[htbp]
  \centering
  \vspace{-6pt}
  \caption{Training configurations of three training stages.}
  \vspace{-6pt}
  \label{training_config}
  \footnotesize  
  \renewcommand\tabcolsep{2pt}  
  \renewcommand\arraystretch{1.1}  
  \begin{tabular}{lccccc}  
    \toprule
    \textbf{Stage} & \textbf{Objective} & \textbf{Epoch} & \textbf{lr} & \textbf{Batch} & \textbf{Param.} \\
    \midrule
    Stage-1 & Alignment & 3 & $5 \times 10^{-5}$ & 512 & - \\
    Stage-2 & SFT & 2 & $3 \times 10^{-5}$ & 64 & - \\
    Stage-3 & GRPO & 1 & $1 \times 10^{-6}$ & 32 & $\beta=0.001$ \\
    \bottomrule
  \end{tabular}
\end{table}
\vspace{-15pt}

\subsection{Performance Analysis}

\noindent\textbf{Main Results.} 
Cross-dataset generalization performance is summarized in Table \ref{tab:dfd_performance}. Our VRAG-DFD framework achieves state-of-the-art results on four out of five evaluation datasets (CDF-v1 \cite{cdf}, CDF-v2 \cite{cdf}, FFIW \cite{ffiw}, and WDF \cite{wdf}). These results strongly demonstrate that our approach leveraging RAG-based dynamic evidence retrieval coupled with critical reasoning significantly enhances the model's ability to generalize to unseen forgery types and complex data distributions.

\noindent\textbf{Analysis on the DFDC Dataset.} 
We observe that the performance on the DFDC dataset does not reach SOTA. This limitation is expected under our \textit{restricted experimental setting}, where the RAG database is exclusively built from FF++ samples. The DFDC dataset contains a large number of samples that differ from FF++ in both visual domain characteristics and forgery artifacts. Consequently, our forensic evidence retriever  fails to retrieve highly relevant historical cases. The external evidence accessible to the MLLM is either of low quality or irrelevant, which interferes with its critical reasoning process and leads to degraded performance. This observation, in fact, validates our central claim: the detection capability of MLLMs is strongly dependent on the quality of the external evidence they can access.

\subsection{Ablation Studies}

\begin{table}[tp]  
    \centering
    \small  
    \renewcommand\arraystretch{1}  
    \renewcommand\tabcolsep{3pt}     
    \vspace{-8pt}
    \caption{{Ablation study of RAG (AUC \%). \textit{w/o} RAG removes the forensic retriever and all associated critical reasoning components.}}
    \vspace{-8pt}
    \label{tab:abla-RAG}
    \begin{tabular}{c|ccccc}  
        \toprule
        \multirow{2}*{Metrics} & \multicolumn{5}{c}{\underline{Test Set AUC}} \\
             & CDF1 & CDF2 & DFDC & FFIW & WDF \\
            \hline
            \textit{w/o} RAG Module & 91.13 & 87.85 & 72.42 & 78.62 & 71.62  \\
            VRAG-DFD & \textbf{99.60} & \textbf{95.97} & \textbf{81.82} & \textbf{93.49} & \textbf{88.96}  \\
        \bottomrule
    \end{tabular}
\end{table}

\begin{table}[tp]
    \centering
    \small
    \renewcommand\arraystretch{1}
    \renewcommand\tabcolsep{3pt}
    \caption{{Ablation study of the three-stage training pipeline. We report AUC (\%) on five generalization benchmarks. }}
    \vspace{-8pt}
    \label{tab:abla-stage}
    \begin{tabular}{c|ccccc}  
        \toprule
        \multirow{2}*{Metrics} & \multicolumn{5}{c}{\underline{Test Set AUC}} \\
        & CDF1 & CDF2 & DFDC & FFIW & WDF \\
        \midrule
        Stage-1 & 96.77 & 94.03 & 78.54 & 90.89 & 87.63  \\
        Stage-1\&2  & 91.53 & \textbf{95.8} & \textbf{81.58} & \textbf{92.37} & \textbf{89.13} \\
        VRAG-DFD & \textbf{99.60} & \textbf{95.97} & \textbf{81.82} & \textbf{93.49} & \textbf{88.96}   \\
        \bottomrule
    \end{tabular}
\end{table}

We conduct a series of comprehensive ablation studies to deconstruct the VRAG-DFD framework and validate the effectiveness of each design choice.

\noindent\textbf{Effectiveness of Forensic RAG.} 
To quantify the contribution of our core RAG-based forensic retrieval component, we evaluate a "\textit{w/o} RAG" variant that relies solely on the MLLM's internal knowledge for detection. As shown in Table~\ref{tab:abla-RAG}, removing the RAG leads to a substantial drop in performance across all datasets, demonstrating that dynamic and targeted external evidence is crucial for robust forensic detection.

\noindent\textbf{Effectiveness of Three-Stage Pipeline.}
Table \ref{tab:abla-stage} reports the ablation results across the three training stages. Using only Stage-1, the model acquires basic visual perception but lacks reasoning ability, leading to limited performance. Incorporating Stage-2 (Critical SFT) yields a substantial improvement, demonstrating the effectiveness of the F-CoT dataset and structured reasoning format (Sec.~3.2.2) in enabling forensic analysis. Further adding Stage-3 (RL-GRPO) enhances critical thinking, indicating that the process-aware reward model (Sec.~3.4.3) promotes a transition from imitation to autonomous decision-making. Overall, the consistent performance gains confirm that each stage is necessary and complementary.

\noindent\textbf{Comparison of Forensic Retrievers.} 
To ensure that the superiority of VRAG-DFD does not rely on a particular state-of-the-art detector (\textit{i.e.}, Effort \cite{effort}),  
We also compare the impact of different retrievers on the final performance. Specifically, we evaluate an Effort-based retriever \cite{effort}, which employs orthogonal subspace decomposition, against a standard CLIP-ViT \cite{clip} retriever fine-tuned with LoRA \cite{lora} on the same dataset.
As shown in Table \ref{tab:retrieval_backbone}, the clip-lora-based retriever achieves superior overall performance. It demonstrates that VRAG-DFD does not depend  
on Effort’s specialized SVD design; instead, a simpler and more generic tuning method (LoRA) unlocks greater potential for forensic retrieval. 


\begin{table}[tp]
    \centering
    \footnotesize
    \renewcommand\arraystretch{0.95}
    \setlength{\tabcolsep}{2.5pt}
    \vspace{-8pt}
    \caption{Ablation study of retriever (AUC \%). ‘CLIP-LoRA’ replaces Effort~\cite{effort} with a standard CLIP model fine-tuned by LoRA.}
    \vspace{-5pt}
    \label{tab:retrieval_backbone}
    \resizebox{\linewidth}{!}{
    \begin{tabular}{c|ccccc|c}
        \toprule
        \multirow{2}*{Retrievers} & \multicolumn{5}{c|}{\underline{Test Set AUC}} & \multirow{2}*{Avg.} \\
        & CDF1 & CDF2 & DFDC & FFIW & WDF & \\
        \midrule
        Effort \cite{effort} & 99.60 & \textbf{95.97} & 81.82 & 93.49 & \textbf{88.96} & 91.62 \\
        \textbf{CLIP-LoRA} \cite{lora} & \textbf{100.00} & 95.77 & \textbf{82.47} & \textbf{94.00} & 88.42 & \textbf{92.13} \\
        \bottomrule
    \end{tabular}
    }
\end{table}

\begin{table}[t]
    \centering
    \caption{Explanation quality evaluation. GPT-4o and Gemini~2.5~Pro serve as evaluators. VRAG-DFD achieves the highest average score.}
    \label{tab:exp_quality_simple}
    \renewcommand{\arraystretch}{1.0}
    \small
    \vspace{-8pt}
    \begin{tabular}{lccc}
        \toprule
        {Model} & {GPT-4o} & {Gemini2.5 Pro} & {Avg.} \\
        \midrule
        GPT-4o & 4.60 & 3.25 & 3.93 \\
        Gemini2.5 Pro & 7.31 & 7.02 & 7.16 \\
        \textbf{VRAG-DFD} & \textbf{7.55} & \textbf{7.78} & \textbf{7.66} \\
        \bottomrule
    \end{tabular}
\end{table}

\noindent\textbf{Evaluation of Explanation Quality} 
To quantitatively evaluate the reliability of model-generated explanations, we conduct a multi-dimensional assessment on 100 randomly selected samples. Explanations are produced by three models: GPT-4o, Gemini 2.5 Pro, and our VRAG-DFD framework. We further employ GPT-4o and Gemini 2.5 Pro as independent evaluators to score each explanation along three standard dimensions (0--3 points each): accuracy, faithfulness, and professionalism. These criteria collectively measure whether the prediction is correct, whether the reasoning is grounded in visual evidence, and whether the explanation follows forensic conventions.
The results in Table~\ref{tab:exp_quality_simple} show that VRAG-DFD substantially outperforms general-purpose MLLMs.




\section{Conclusion}




To achieve the high-quality associated forgery knowledge and critical reasoning abilities in MLLM-based DFD detection, we firstly constructed two datasets: Forensic Knowledge Database (FKD) and Forensic Chain-of-Thought Dataset (F-CoT). 
For critical CoT construction, we propose the VRAG-DFD with accurate dynamic forgery knowledge retrieval and powerful critical reasoning capabilities. VRAG-DFD achieved SOTA and competitive performance on DFD generalization benchmarks.
This research confirms the application prospects of RAG in the field of DFD.

{
    \small
    \bibliographystyle{ieeenat_fullname}
    \bibliography{main}
}

\clearpage
\appendix
\clearpage
\setcounter{page}{1}
\maketitlesupplementary

\section{Implementation Details}

\subsection{Retrieval and Inference Settings}
In our dynamic forensic retrieval module, we set the number of retrieved evidence items to $k=5$. This choice is based on empirical observations on the validation set, where $k=5$ provides an effective trade-off between supplying sufficient supportive evidence and maintaining efficient MLLM reasoning. Larger values of $k$ tend to introduce unnecessary noise due to excessively long context windows, while smaller values lead to insufficient contextual grounding.

\subsection{Training Configuration and Data Augmentation}
As described in the main paper, the training pipeline of VRAG-DFD consists of three stages. To maximize model performance, we adopt tailored training strategies for each stage. 

\noindent\textbf{Stage-1: Foundational Visual Alignment.}
We use approximately 80,000 images from the FaceForensics++ (FF++) training set. 

\begin{itemize}
    \item \textbf{Training Strategy.} This stage aims to establish the model’s foundational visual sensitivity to forgery cues. We therefore perform LoRA fine-tuning on the ViT (vision encoder), the Aligner (projection module), and the LLM.
    \item \textbf{Data Augmentation.} To enhance the visual encoder’s robustness and mitigate overfitting, we apply a set of augmentations implemented using the Albumentations library, such as Horizontal Flip, Image Compression, HueSaturationValue Adjustment.
\end{itemize}    

\noindent\textbf{Stage-2 \& Stage-3: Critical Reasoning Training.}
Both Stage-2 and Stage-3 are trained on 9,000 samples sampled from the FF++ dataset. In Stage-2, these samples are used for supervised fine-tuning (SFT) with F-CoT annotations.

\begin{itemize}
    \item \textbf{Training Strategy.} During these stages, we freeze all ViT parameters and only lora finetune the Aligner and LLM. This design preserves the general visual features learned in Stage-1, while encouraging the model to focus on logical reasoning, evidence aggregation, and conflict resolution.
    \item \textbf{Data Augmentation.} To ensure that the model can capture extremely subtle pixel-level forgery traces, no data augmentation is applied in Stage~2 and Stage~3. This avoids introducing artificial distortions that could obscure genuine forgery artifacts.
\end{itemize}

\section{Database Construction Details}

To support the VRAG-DFD framework, we construct two core datasets:  
(1) the \textit{Forensic Knowledge Database (FKD)} for retrieval, and  
(2) the \textit{Forensic Chain-of-Thought (F-CoT)} dataset for critical reasoning training.

\subsection{Construction of the Forensic Knowledge Database (FKD)}

\paragraph{Data Sources and Balancing Strategy.}
The original FF++ dataset exhibits a Real:Fake ratio of approximately 1:4. To build a category-balanced retrieval database, we adopt the following sampling strategy, which produces a balanced 1:1 distribution of real and manipulated samples:

\begin{itemize}
    \item \textbf{Fake Frames:} From each video of the four manipulation types (DeepFakes, Face2Face, FaceSwap, NeuralTextures), we randomly sample around 3 frames, resulting in approximately 8,000 fake frames.
    \item \textbf{Real Frames:} From each real video, we sample approximately 12 frames, also yielding around 8,000 real frames.
\end{itemize}

\paragraph{Expert-Level Annotation.}
To ensure high-quality and professional annotations, we adopt a combination of contrastive annotation and a structured forgery guideline. Using Gemini 2.5 Pro, we input each target image together with its corresponding real/fake counterpart and instruct the model to produce detailed descriptions of forged regions, guided by manipulation-specific characteristics (e.g., texture discontinuities in NeuralTextures). The full prompt is shown in the END of this supplementary material.

\subsection{Construction of the Forensic Chain-of-Thought (F-CoT) Dataset}

\paragraph{Data Source.}
We select 1,000 videos from the FF++ training split that are not used during Stage~1 training. From each video, 9 frames are sampled, resulting in a dataset of 9,000 samples.

\paragraph{Reference Information Generation.}
For each sample, we retrieve the top-$k$ most relevant images from FKD (excluding itself) and use their text annotations as external reference information.

\paragraph{Critical Logic Construction.}
To encourage the model to reason critically about retrieved evidence, we categorize each sample into one of three types based on the Stage~1 prediction and the quality of retrieved evidence. Gemini 2.5 Pro, acting as a “teacher” with access to ground-truth labels, is prompted to produce corresponding chain-of-thought reasoning:

\begin{itemize}
    \item \textbf{Type-1: Cross-Validation.}  
    \textit{Condition:} Stage~1 predicts correctly.  
    \textit{Objective:} Teach the model to verify its visual judgment using supportive retrieved evidence while ignoring irrelevant noise.

    \item \textbf{Type-2: Evidence-Guided Correction.}  
    \textit{Condition:} Stage~1 predicts incorrectly, but retrieved evidence contains strong, reliable cues.  
    \textit{Objective:} Guide the model to recognize its initial mistake and rely on high-quality external evidence for correction.

    \item \textbf{Type-3: Resilient Rejection.}  
    \textit{Condition:} Stage~1 predicts incorrectly, and retrieved evidence is misleading (e.g., semantically similar but different manipulation type).  
    \textit{Objective:} Train the model to identify unreliable evidence, reject misleading cues, and fall back to its internal visual knowledge for the final judgment.
\end{itemize}

The detailed F-CoT generation prompt in the END of this supplementary material.

\section{Additional Experimental Results}

\subsection{Data Quality Comparison: FKD vs. DDVQA.}
To further examine how knowledge quality influences retrieval-based reasoning, we replace the FKD retrieval corpus in VRAG-DFD with the textual descriptions from the DDVQA \cite{ddvqa} dataset and conduct experiments under identical settings.
As shown in Table~\ref{tab:data_quality}, replacing FKD with DDVQA results in a substantial performance drop.
This observation demonstrates that the fine-grained, expert-level forensic annotations contained in FKD are crucial for enhancing reasoning ability.
In contrast, the descriptions in DDVQA are relatively coarse and lack professional forensic knowledge, making them insufficient for providing stable and reliable forgery cues.
Overall, these findings underscore the critical importance of high-quality forensic knowledge for cross-dataset generalization and robust reasoning.

\begin{table}[tp]
    \centering
    \small
    \renewcommand\arraystretch{1}
    \renewcommand\tabcolsep{3pt}
    \caption{\textbf{Data Quality Comparison between FKD and DDVQA.} 
Replacing FKD with DDVQA leads to a significant performance drop, 
indicating that forensic-grade knowledge is essential for effective reasoning.}
    \label{tab:data_quality}
    \resizebox{\linewidth}{!}{
    \begin{tabular}{c|ccccc|c}  
        \toprule
        \multirow{2}*{Metrics} & \multicolumn{5}{c|}{\underline{Test Set AUC}} & \multirow{2}*{Avg.} \\
        & CDF1 & CDF2 & DFDC & FFIW & WDF &\\
        \midrule
        DD-VQA~\cite{ddvqa} & 96.17 & 94.83 & 79.48 & 92.50 & 85.88 & 89.77 \\
        \textbf{FKD (Ours)} & \textbf{99.60} & \textbf{95.97} & \textbf{81.82} & \textbf{93.49} & \textbf{88.96} & \textbf{91.62}  \\
        \bottomrule
    \end{tabular}
    }
\end{table}

\subsection{Contribution of the RAG Module Across Stages}

To quantify the efficiency of external knowledge retrieval, we conduct an ablation study by selectively activating the RAG module at each training stage. The results in Table~\ref{tab:ablation_rag_stages} demonstrate substantial and consistent gains across all stages. This confirms that the RAG mechanism is an integral component of our framework, consistently bridging the knowledge gap from initial visual alignment to final critical reasoning.

\begin{table}[t]
    \centering
    \caption{\textbf{Ablation study on the impact of the RAG module across three training stages.} We report the AUC (\%) on five generalization benchmarks. The column `$\Delta$ Avg.' highlights the performance gain attributable to the RAG module within each stage. Crucially, we observe that RAG provides \textbf{substantial improvements consistently across all stages}, demonstrating that dynamic retrieval of expert knowledge is fundamental to bridging the generalization gap, even in the early visual alignment phase.}
    \label{tab:ablation_rag_stages}
    \resizebox{\linewidth}{!}{
    \begin{tabular}{c|c|ccccc|c}
    \toprule
    \multirow{2}*{Stage} & \multirow{2}*{Config} & \multicolumn{5}{c|}{\underline{Test Set AUC}} & \multirow{2}*{$\Delta$} \\
     & & {CDF1} & {CDF2} & {DFDC} & {FFIW} & {WDF} &  \\
    \midrule
    
    \multirow{2}{*}{\textbf{Stage-1}}
    & \textit{w/o} RAG 
    & 89.72 & 86.30 & 73.31 & 78.59 & 78.73 
    & \multirow{2}{*}{\textbf{\textcolor{red}{+8.16}}} \\
    
    & \textbf{w/ RAG}
    & 96.77 & 94.03 & 78.54 & 90.89 & 87.63 
    & \\
    
    \midrule
    
    \multirow{2}{*}{\textbf{Stage-2}}
    & \textit{w/o} RAG 
    & 91.33 & 86.96 & 71.57 & 76.43 & 70.05 
    & \multirow{2}{*}{\textbf{\textcolor{red}{+10.66}}} \\
    
    & \textbf{\textit{w/} RAG}
    & 91.53 & 95.80 & 81.58 & 92.37 & 89.13 
    & \\
    
    \midrule
    
    \multirow{2}{*}{\textbf{Stage-3}}
    & \textit{w/o} RAG 
    & 91.53 & 86.98 & 72.05 & 76.51 & 70.04 
    & \multirow{2}{*}{\textbf{\textcolor{red}{+12.2}}} \\
    
    & \textbf{\textit{w/} RAG}
    & 99.6 & 95.97 & 81.82 & 93.49 & 88.96 
    & \\
    
    \bottomrule
    \end{tabular}
    }
\end{table}

\subsection{Effectiveness on Different Base MLLMs.} 
To verify the model-agnostic nature of the VRAG-DFD framework, we replace the default \texttt{Qwen2.5VL} with \texttt{InternVL3-8B}, applying our full three-stage training pipeline to each. As summarized in Table \ref{tab:abla-MLLM}, all base models exhibit substantial improvements in cross-dataset detection performance after training under our framework. This demonstrates that our approach serves as a general and effective paradigm for teaching MLLMs advanced forensic reasoning skills.
\begin{table}[tp]
    \centering
    \small
    \renewcommand\arraystretch{1}
    \renewcommand\tabcolsep{1pt}
    \caption{{Effectiveness across different base MLLMs (AUC \%). 
        We show the performance of each model before (zero-shot) and after ('+ VRAG-DFD (Ours)') applying our three-stage VRAG-DFD pipeline. }}
    \vspace{-8pt}
    \label{tab:abla-MLLM}
    \begin{tabular}{c|ccccc}  
        \toprule
        \multirow{2}*{Metrics} & \multicolumn{5}{c}{\underline{Test Set AUC}} \\
            &  CDF1 & CDF2 & DFDC & FFIW & WDF \\
            \hline
            Qwen2.5VL-7B & 66.00 & 73.35 & 56.82 & 71.91 & 66.82  \\
            Qwen2.5VL-7B + ours & \textbf{99.60} & \textbf{95.97} & \textbf{81.82} & \textbf{93.49} & \textbf{88.96}  \\
            \midrule
            Intern3VL-8B & 66.33 & 72.06 & 57.19 & 70.86 & 65.75  \\
            Intern3VL-8B + ours & \textbf{99.19} & \textbf{95.09} & \textbf{80.16} & \textbf{92.45} & \textbf{90.00} \\
        \bottomrule
    \end{tabular}
\end{table}

\subsection{Latency Analysis of the RAG Module}
The construction and indexing of the FKD database are performed entirely offline, ensuring that no additional inference overhead is introduced during deployment. During inference, the RAG module only performs FAISS-based retrieval over precomputed feature embeddings. As shown in Tab.~\ref{tab:flops}, the retrieval operation accounts for only 0.35\% of the total computational cost. This negligible overhead indicates that the integration of the RAG module has a minimal impact on overall inference latency.

\begin{table}[h]
\centering
\caption{Computational complexity analysis of VRAG-DFD.}
\label{tab:flops}
\scriptsize
\renewcommand{\arraystretch}{0.75}
\setlength{\tabcolsep}{3.5pt}
\begin{tabular}{lcccccc}
\toprule
 & \multicolumn{2}{c}{Retrieval} & \multicolumn{2}{c}{Inference} & \multicolumn{2}{c}{Total} \\
\cmidrule(lr){2-3} \cmidrule(lr){4-5} \cmidrule(lr){6-7}
Metric & GFLOPs & Ratio & GFLOPs & Ratio & GFLOPs & Ratio \\
\midrule
Value & $\sim$81 & \textbf{0.35\%} & $\sim$22,760 & 99.65\% & $\sim$22,841 & 100.00\% \\
\bottomrule
\end{tabular}
\vspace{-4pt}
\end{table}

\subsection{Impact of Frame Sampling Density in Stage-1}

In Stage-1, we adopt dense frame sampling over all 2,500 training videos (approximately 80k images) to maximize the model's visual alignment capability. 
To validate the necessity of this computationally intensive strategy, we compare it with sparse sampling, where only 8 frames per video are selected.

\textbf{Results.} As reported in Table~\ref{tab:ablation_stage1_sampling}, dense sampling significantly outperforms sparse sampling, increasing the average AUC from 77.61\% to 81.41\%, corresponding to a performance gain of nearly +3.8\%.

\textbf{Analysis.} This substantial improvement reveals two key insights:

\begin{enumerate}
    \item \textbf{Necessity of Dense Supervision.} Forgery artifacts in deepfakes, such as frame-to-frame jitter and transient texture anomalies, are often sparse and non-continuous. Sparse sampling may miss these critical discriminative frames, leading to an incomplete feature space.
    \item \textbf{Solid Foundation for Subsequent Stages.} Stage-1 visual alignment serves as the cornerstone of the entire VRAG-DFD framework. Our experiments indicate that dense training with all available frames is essential to obtain a sufficiently robust visual encoder, which underpins the reasoning-centric Stage-2 and Stage-3.
\end{enumerate}

\begin{table}[t]
    \centering
    \caption{\textbf{Ablation study on frame sampling density in Stage-1.} We compare sparse sampling (8 frames/video) against dense sampling (all frames). The results demonstrate a \textbf{substantial performance leap} in the average AUC (+3.80\%), confirming that dense visual supervision is indispensable for learning robust forgery features across diverse datasets.}
    \label{tab:ablation_stage1_sampling}
    \resizebox{\linewidth}{!}{
    \begin{tabular}{l|c|ccccc|c}
    \toprule
    \multirow{2}*{Samp. Strategy} & \multirow{2}*{Total} & \multicolumn{5}{c|}{\underline{Test Set AUC}} & \multirow{2}*{Avg.} \\
     &  & {CDF1} & {CDF2} & {DFDC} & {FFIW} & {WDF} & \\
    \midrule
    Sparse (8 fms) & $\sim$28k & \textbf{92.14} & 76.85 & 70.69 & 70.40 & 77.95 & 77.61 \\
    \textbf{Dense (All fms)} & \textbf{$\sim$80k} & 89.72 & \textbf{86.30} & \textbf{73.71} & \textbf{78.59} & \textbf{78.73} & \textbf{81.41} \\
    \midrule
    {{Gain ($\Delta$)}} & - &\textit{-2.42} & \textbf{\textcolor{red}{+9.45}} & \textbf{\textcolor{red}{+3.02}} & \textbf{\textcolor{red}{+8.19}} & \textbf{\textcolor{red}{+0.78}} & \textbf{\textcolor{red}{+3.80}} \\
    \bottomrule
    \end{tabular}
    }
\end{table}

\begin{table}[t]
    \centering
    \caption{\textbf{Robustness against misleading retrieval evidence.} We evaluate a subset of adversarial samples where the initial visual perception is correct ($s_1$=Correct), but the RAG evidence is misleading ($RAG$=False). The results demonstrate that VRAG-DFD maintains an exceptionally high \textbf{Robustness Rate} ($>96\%$ on avg.), effectively rejecting misleading external noise.}
    \label{tab:robustness_rag_errors}
    \setlength{\tabcolsep}{2pt}
    \scriptsize
    \begin{tabular}{@{}l|ccccc|c@{}}
    \toprule
    \multirow{2}*{Metric} & \multicolumn{5}{c|}{\underline{Test Set AUC}} & \multirow{2}*{Weigh. Avg.} \\
     & {CDF-v1} & {CDF-v2} & {DFDC} & {FFIW} & {WDF} & \\
    \midrule
    \# Adv. Samp. & 34 & 429 & 534 & 1,374 & 5,226 & 7,597 \\
    \# VRAG\_DFD Cor. & 32 & 417 & 520 & 1,345 & 5,048 & 7,362 \\
    \midrule
    \textbf{Robust. Rate (\%)} & 94.12 & 97.20 & 97.38 & 97.89 & 96.59 & \textbf{\textcolor{red}{96.91}} \\
    \bottomrule
    \end{tabular}
\end{table}
\subsection{Robustness against Adversarial Retrieval Noise}

To evaluate whether VRAG-DFD can resist misleading external evidence, we specifically analyze a \emph{difficult subset} of the test set. This subset is defined as samples where the retriever fails, i.e., the majority of the top-$k$ retrieved evidences belong to incorrect categories.

\textbf{Results.} As shown in Table~\ref{tab:robustness_rag_errors}, VRAG-DFD demonstrates remarkable resilience even under strongly misleading retrieval conditions. Across all five evaluation datasets, the proportion of correctly classified samples consistently exceeds 94\%.

\textbf{Analysis.} These results indicate that VRAG-DFD does not simply ``copy'' the retrieved evidence. Instead, it successfully learns to critically weigh the reliability of external evidence and perform reasoning that accounts for potential inconsistencies, highlighting its robust and principled inference capabilities.

\subsection{Qualitative Analysis}
\begin{figure*}[htb] 
  \centering  
  \vspace{6pt}  
  \includegraphics[width=1.0\textwidth]{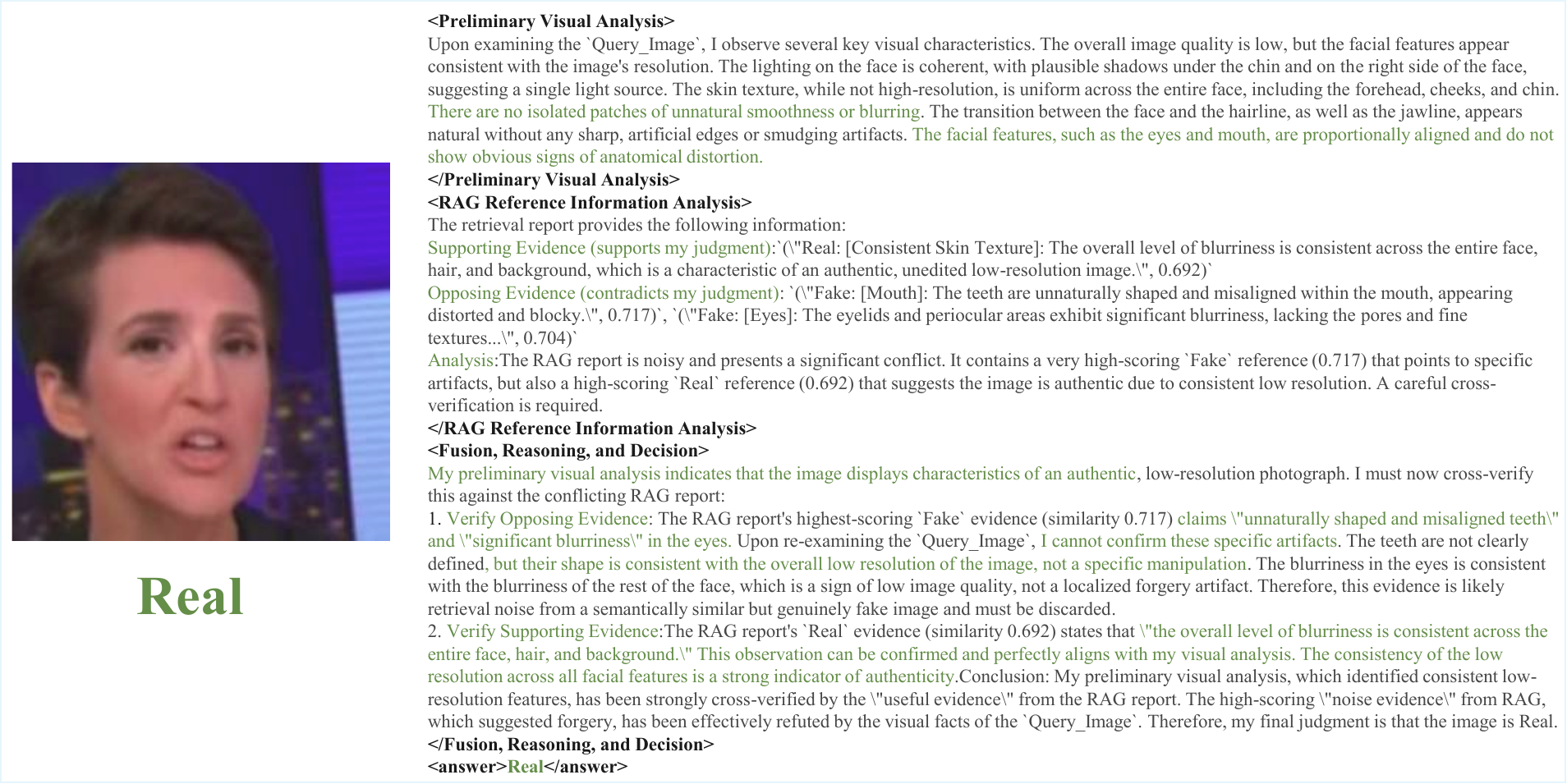}  
  \vspace{4pt}  
  \caption{Qualitative analysis of VRAG-DFD (real case 1).}  
  \label{case1}  
  \vspace{8pt}  
  \includegraphics[width=1.0\textwidth]{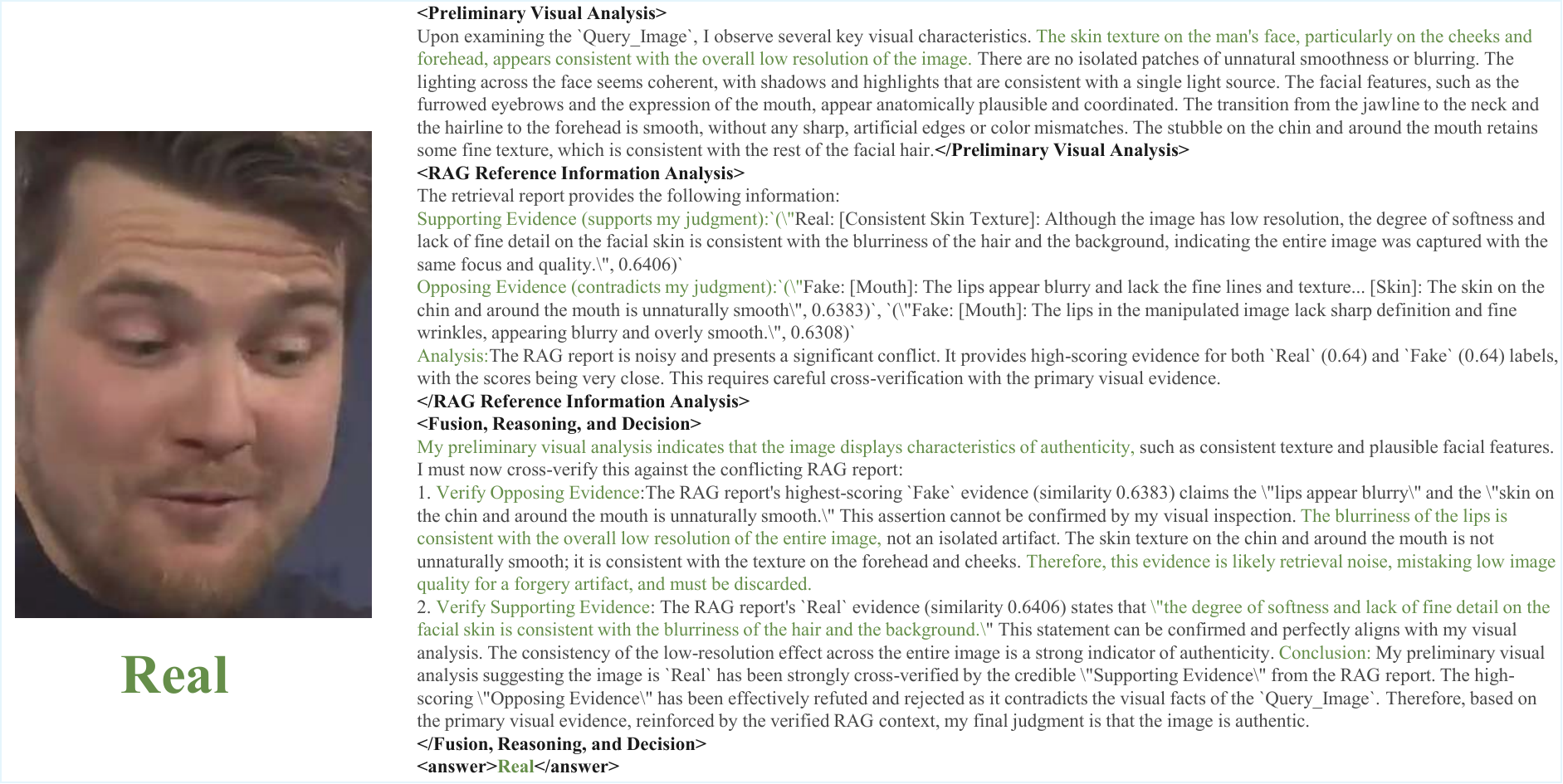}  
  \vspace{4pt}
  \caption{Qualitative analysis of VRAG-DFD (real case 2).}  
  \label{case2}  
  \vspace{6pt}  
\end{figure*}

\begin{figure*}[htb] 
  \centering  
  \vspace{6pt}  
  \includegraphics[width=1.0\textwidth]{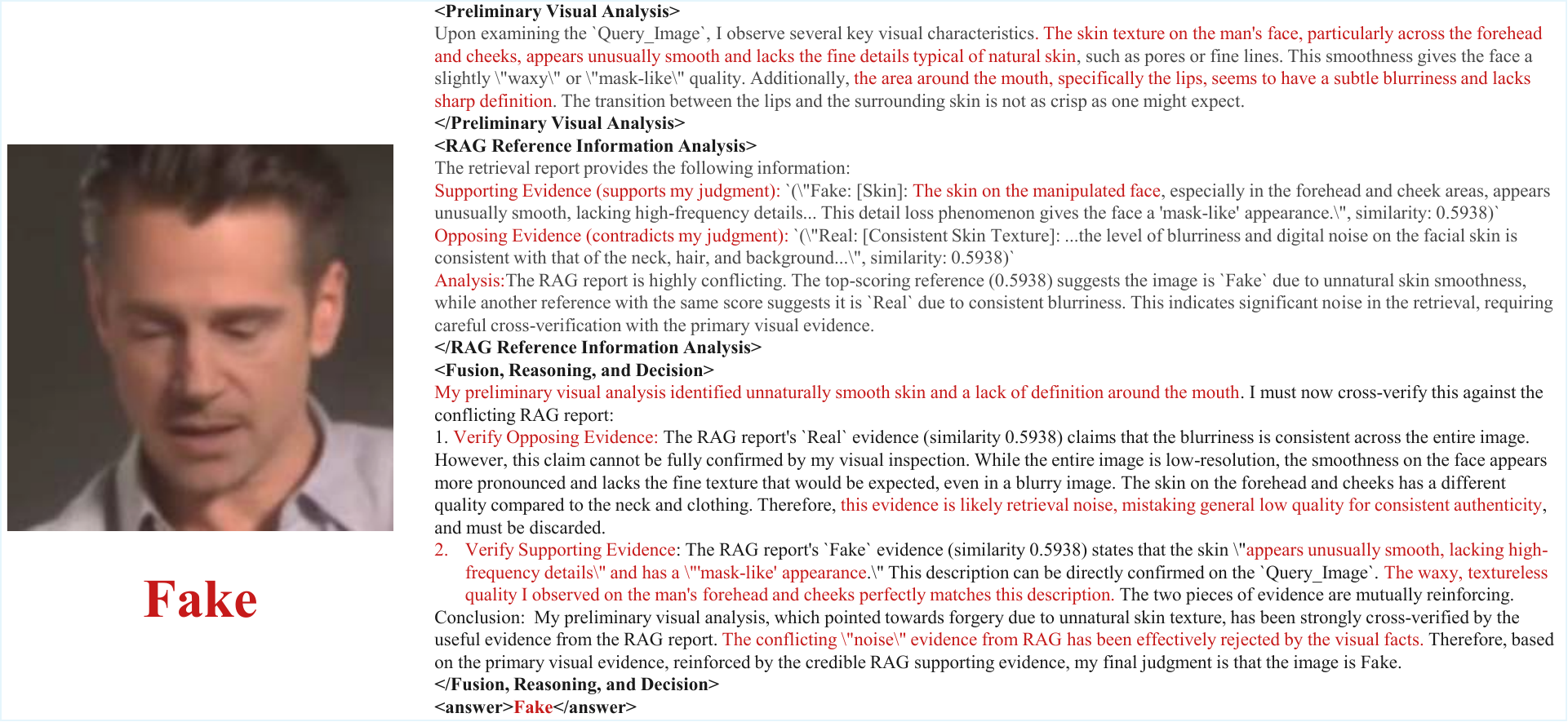}  
  \vspace{4pt}  
  \caption{Qualitative analysis of VRAG-DFD (fake case 1).}  
  \label{case3}  
  \vspace{8pt}  
  \includegraphics[width=1.0\textwidth]{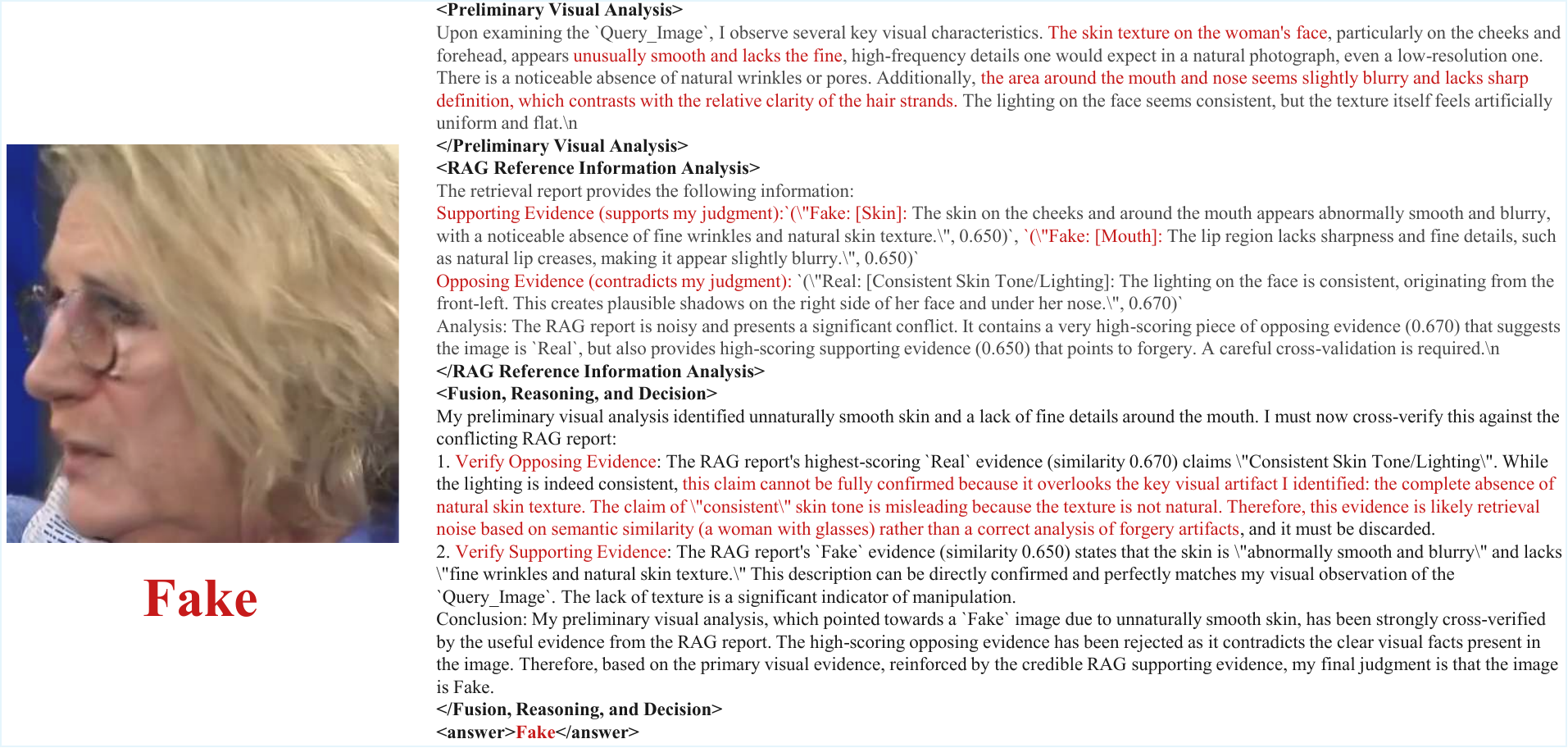}  
  \vspace{4pt}
  \caption{Qualitative analysis of VRAG-DFD (fake case 2).}  
  \label{case4}  
  \vspace{6pt}  
\end{figure*}

To visually demonstrate the interpretability and reasoning capabilities of VRAG-DFD, we present qualitative examples in Figures \ref{case1}, \ref{case2}, \ref{case3} and \ref{case4}. These figures display the query images alongside the complete Forensic Chain-of-Thought (F-CoT) responses generated by our model, illustrating how it integrates visual analysis and retrieval evidence to reach a final verdict.

\newpage

\newpage

\newpage

\begin{figure*}[t] 
    
    \begin{tcolorbox}[
        colback=gray!5,
        colframe=black,
        title={\textbf{Prompt for DeepFakes Annotation}},
        rounded corners, 
        width=\textwidth,
        enlarge left by=0mm,
        enlarge right by=0mm,
        boxrule=0.8pt
    ]
    
    \textbf{System Prompt:} \\
    You are an expert in face tampering detection. Your task is to strictly compare the \texttt{[Manipulated Image]} and the \texttt{[Original Image]}, first precisely locate the manipulated regions, and then \textbf{explain} how the forgery artifacts within those regions were caused by the improper processing of DeepFakes technology. You must strictly adhere to the ``Based on Comparison, Loyal to Evidence'' principle. \textbf{Hallucinations are strictly forbidden}. Focus on artifacts left by the \textbf{forgery process} (e.g., blending, alignment), not natural facial features (e.g., makeup, appearance). Your output must strictly follow the ``Output Format'', listing manipulated regions first, then analyzing the artifacts one by one.
    
    \vspace{2mm}
    \textbf{User Prompt:} \\
    \textbf{Task Definition:} You will receive two images: the first is a \texttt{[Manipulated Image]} generated using DeepFakes technology, and the second is the corresponding \texttt{[Original Image]}. Carefully compare the two images to locate manipulated regions, then explain the forgery artifacts using the ``DeepFakes Potential Forgery Artifacts Reference Guide''. Focus solely on artifacts caused by DeepFakes, not natural differences.
    
    \vspace{1mm}
    \textbf{Core Analysis Principles:}
    \begin{enumerate}
        \item Carefully compare all facial features, hair, skin color, etc., to locate manipulated regions without omissions or hallucinations.
        \item Distinguish between:
        \begin{itemize}
            \item \textbf{Analyze This - Forgery Artifacts}: Anomalies caused by technical flaws (e.g., low-resolution bottlenecks, blending, alignment failures).\\
            \textit{Correct Example}: Inconsistent skin tone, feature overlap, blurry edges, structural distortion.
            \item \textbf{Ignore This - Natural Features}: Inherent facial features unrelated to forgery.\\
            \textit{Incorrect Example}: Makeup, moles, dimples, natural face shape or expression.
        \end{itemize}
        \item Attribute observed phenomena (e.g., ``skin looks airbrushed'') to the technical flaw (e.g., ``detail loss due to low resolution'').
        \item Report all types of artifacts present in a region, not only the most obvious.
    \end{enumerate}
    
    \vspace{1mm}
    \textbf{DeepFakes Potential Forgery Artifacts Reference Guide:}
    \begin{enumerate}
        \item \textbf{Blending Border Artifacts (High-Incident Area)}
        \begin{itemize}
            \item \textit{Cause}: Poisson blending at facial edges (face periphery, hair, chin, neck).
            \item \textit{Manifestation}: Edge color differences, unnatural blurring/halo effects along contours.
        \end{itemize}
        \item \textbf{Structural Abnormality (Feature Misalignment/Distortion)}
        \begin{itemize}
            \item \textit{Cause}: Keypoint alignment failure.
            \item \textit{Manifestation}: Distorted or misaligned facial features; feature overlap (e.g., eyebrows).
        \end{itemize}
        \item \textbf{Detail Loss}
        \begin{itemize}
            \item \textit{Cause}: Low-resolution face generation.
            \item \textit{Manifestation}: Blurry eyes, teeth, irises; abnormally smooth skin lacking texture or pores.
        \end{itemize}
    \end{enumerate}
    
    \vspace{1mm}
    \textbf{Output Format:} \\
    First list manipulated regions, then describe the corresponding forgery artifacts.\\
    \textit{Reference Example:} \\
    \textit{Manipulated Regions: Skin, Eyebrows\\
    Forgery Artifacts: [Skin]: Central area cool white, periphery yellowish-black, clear blending boundary. [Eyebrows]: Ghosting caused by facial alignment failure.}\\
    \\
    \textbf{Here are \texttt{[Manipulated Image]} and the \texttt{[Original Image]}.}\\
    \texttt{[Manipulated Image]}: \{\{manipulated\_image\_path\}\} \\
    \texttt{[Original Image]}: \{\{original\_image\_path\}\}
    
    \end{tcolorbox}
    
\end{figure*}

    
    
    

\begin{figure*}[t] 
    
    \begin{tcolorbox}[
        colback=gray!5,
        colframe=black,
        title={\textbf{Prompt for Face2Face Annotation}},
        rounded corners, 
        width=\textwidth,
        enlarge left by=0mm,
        enlarge right by=0mm,
        boxrule=0.8pt
    ]
    
    \textbf{System Prompt:} \\
    You are an expert in face tampering detection. Your task is to strictly compare the \texttt{[Manipulated Image]} and the \texttt{[Original Image]}, first precisely locate the manipulated regions, and then \textbf{explain} how the forgery artifacts within those regions were \textbf{caused by the Face2Face 3D model re-rendering process}. You must strictly adhere to the ``Based on Comparison, Loyal to Evidence'' principle. \textbf{Hallucinations are strictly forbidden}. Focus on artifacts left by the \textbf{forgery process} (e.g., 3D model tracking, re-rendering, expression transfer), not natural facial features (e.g., makeup, appearance). Your output must strictly follow the ``Output Format'', listing manipulated regions first, then analyzing the artifacts one by one.
    
    \vspace{2mm}
    \textbf{User Prompt:} \\
    \textbf{Task Definition:} You will receive two images: the first is a \texttt{[Manipulated Image]} generated using Face2Face technology, and the second is the corresponding \texttt{[Original Image]}. Carefully compare the two images to locate manipulated regions, then explain the forgery artifacts using the ``Face2Face Potential Forgery Artifacts Reference Guide''. Focus solely on artifacts caused by Face2Face technology, not natural expression differences.
    
    \vspace{1mm}
    \textbf{Core Analysis Principles:}
    \begin{enumerate}
        \item Identify all differences to locate manipulated areas, including facial features, hair, and skin tone. Pay special attention to the lips and surrounding areas.
        \item Distinguish between:
        \begin{itemize}
            \item \textbf{Analyze This - Forgery Artifacts}: Technical anomalies (3D re-rendering, Blendshape parameterized driving, 3D tracking errors).\\
            \textit{Correct Example}: 'Plastic-like' skin, expression structural distortion ('puppet-like'), misalignment at facial edges, unrealistic lighting/highlights.
            \item \textbf{Ignore This - Natural Features}: Expression changes themselves are not artifacts.\\
            \textit{Incorrect Example}: Different expressions (smiling vs laughing), mouth open, [Manipulated Image] laughing while [Original Image] is smiling.
        \end{itemize}
        \item Base your analysis on the technical flaws outlined in the Reference Guide; attribute observed phenomena to the correct cause.
        \item Note: Face2Face preserves identity; focus on expression manipulation artifacts, not identity differences.
    \end{enumerate}
    
    \vspace{1mm}
    \textbf{Face2Face Potential Forgery Artifacts Reference Guide:}
    \begin{enumerate}
        \item \textbf{3D Model Render/Blend Borders}
        \begin{itemize}
            \item \textit{Cause}: Re-rendered facial regions pasted onto the original frame.
            \item \textit{Manifestation}: Facial edge mismatch, background/hair distortion adjacent to the face.
        \end{itemize}
        \item \textbf{Texture Mismatch Due to Re-rendering}
        \begin{itemize}
            \item \textit{Cause}: 3D model cannot perfectly replicate original camera imaging details or skin textures.
            \item \textit{Manifestation}: 'Plastic-like' skin, unrealistic lighting/highlights.
        \end{itemize}
        \item \textbf{Local Artifact: Mouth Detail Blurring}
        \begin{itemize}
            \item \textit{Cause}: Difficulty preserving fine lip textures during expression manipulation.
            \item \textit{Manifestation}: Loss of lip texture, blurry lip borders, stiff lip shape.
        \end{itemize}
        \item \textbf{Expression Structural Distortion ("Puppet-like" Artifacts)}
        \begin{itemize}
            \item \textit{Cause}: Parameterized Blendshape coefficients limit realistic muscle coordination.
            \item \textit{Manifestation}: Non-ergonomic stretching, stiff/mechanical expressions, lack of natural coordination between mouth and eyes.
        \end{itemize}
    \end{enumerate}
    
    \vspace{1mm}
    \textbf{Output Format:}\\
    First list manipulated regions, then describe corresponding forgery artifacts.\\
    \textit{Reference Example:} \\
    \textit{Manipulated Regions: Facial Contour, Mouth\\
    Forgery Artifacts: [Facial Contour]: Blending artifacts at the contour. [Mouth]: Smile shape stiff and unnatural; corners stretch mechanically.} \\
    \\
    \textbf{Here are \texttt{[Manipulated Image]} and the \texttt{[Original Image]}.}\\
    \texttt{[Manipulated Image]}: \{\{manipulated\_image\_path\}\} \\
    \texttt{[Original Image]}: \{\{original\_image\_path\}\}
    
    \end{tcolorbox}
\end{figure*}
\begin{figure*}[t]
    \centering
    \begin{tcolorbox}[
        colback=gray!5,
        colframe=black,
        title={\textbf{Prompt for FaceSwap Annotation}}, 
        arc=2mm, 
        width=\linewidth, 
        boxrule=0.8pt,
        fontupper=\small, 
        top=2mm, bottom=2mm, left=2mm, right=2mm 
    ]

    \textbf{System Prompt:} \\
    You are an expert in face tampering detection. Your task is to strictly compare the \texttt{[Manipulated Image]} and the \texttt{[Original Image]}, first precisely locate the manipulated regions, and then \textbf{explain} how the forgery artifacts within those regions were \textbf{caused by the improper processing of FaceSwap technology}. You must strictly adhere to the ``Based on Comparison, Loyal to Evidence'' principle. \textbf{Hallucinations are strictly forbidden}. Your analysis of the manipulated regions must focus on the artifacts left by the \textbf{forgery process} (e.g., blending, alignment), not the natural features of the face (e.g., makeup, appearance). Your output must \textbf{strictly follow} the ``Output Format'' specified by the user, listing the manipulated regions first, then analyzing the artifacts one by one.

    \vspace{2mm}
    \textbf{User Prompt:} \\
    \textbf{Task Definition:} You will receive two images, the first is a \texttt{[Manipulated Image]} generated using FaceSwap technology, and the second is the corresponding \texttt{[Original Image]}. Please act as an expert in face tampering detection. By carefully comparing the two images, first find the manipulated regions in the \texttt{[Manipulated Image]}, and then \textbf{explain} the forgery artifacts in those regions. You must first locate the manipulated regions by careful comparison, and then use the ``FaceSwap Potential Forgery Artifacts Reference Guide'' to \textbf{explain} how the artifacts you found were caused by FaceSwap processing. Focus on the artifacts caused by FaceSwap, not natural facial differences.

    \vspace{1mm}
    \textbf{Core Analysis Principles:}
    \begin{enumerate}[leftmargin=*, noitemsep, topsep=0pt]
        \item Carefully compare each facial feature, hair, skin, etc., to locate manipulated regions without omissions or hallucinations.
        \item Distinguish clearly between:
        \begin{itemize}[leftmargin=*, noitemsep, topsep=0pt]
            \item \textbf{Analyze This - Forgery Artifacts}: Visual anomalies directly caused by technical flaws (e.g., blending, alignment, 3D fitting).\\
            \textit{Correct Example}: Inconsistent skin tone, feature overlap, blurry edges, structural distortion.
            \item \textbf{Ignore This - Natural Features}: Inherent facial features unrelated to forgery.\\
            \textit{Incorrect Example}: Makeup, moles, dimples, natural face shape or expression.
        \end{itemize}
        \item Attribute observed phenomena (e.g., ``two eyebrows'') to technical flaws (e.g., ``feature overlap caused by keypoint mismatch'').
    \end{enumerate}

    \vspace{1mm}
    \textbf{FaceSwap Potential Forgery Artifacts Reference Guide:}
    \begin{enumerate}[leftmargin=*, noitemsep, topsep=0pt]
        \item Color \& Lighting Inconsistency
        \begin{itemize}[leftmargin=*, noitemsep]
            \item \textbf{Inconsistent Skin Tone}: Skin hue, saturation, or brightness differs from surrounding regions.
            \item \textbf{Incorrect Lighting/Shadows}: Face highlights/shadows inconsistent with environment.
        \end{itemize}
        \item Feature Misalignment
        \begin{itemize}[leftmargin=*, noitemsep]
            \item \textbf{Facial Structure Abnormality}: Misaligned facial features due to sparse keypoint fitting.
            \item \textbf{Feature Overlap}: Overlapping features from both faces (e.g., eyebrows).
        \end{itemize}
        \item Blending Artifacts
        \begin{itemize}[leftmargin=*, noitemsep]
            \item \textbf{Unnatural Blurring}: Intentional blur to hide splicing lines.
            \item \textbf{Edge Overlap/Gaps}: Misalignment with head contour causing gaps or overlaps.
            \item \textbf{Hard Edges}: Clear cutting sensation from poor blending.
        \end{itemize}
        \item Mask-like Feel
        \begin{itemize}[leftmargin=*, noitemsep]
            \item \textbf{Detail Loss}: Overly smooth/blurry skin creating a mask-like appearance.
        \end{itemize}
    \end{enumerate}

    \vspace{1mm}
    \textbf{Output Format:} \\
    First list the manipulated regions, then describe the corresponding forgery artifacts.\\
    \textit{Reference Example:} \\
    \textit{Manipulated Regions: Eyebrows, Skin \\
    Forgery Artifacts: [Eyebrows]: Ghosting due to \textbf{feature overlap}. [Skin]: Cheek tone mismatch exposing \textbf{blending defect}.}
    
    \vspace{2mm}
    \textbf{Here are \texttt{[Manipulated Image]} and the \texttt{[Original Image]}:}\\
    \texttt{[Manipulated Image]}: \texttt{\{\{manipulated\_image\_path\}\}} \\
    \texttt{[Original Image]}: \texttt{\{\{original\_image\_path\}\}}

    \end{tcolorbox}
    \label{fig:faceswap_prompt}
\end{figure*}

\begin{figure*}[t] 
    
    \begin{tcolorbox}[
        colback=gray!5,
        colframe=black,
        title={\textbf{Prompt for NeuralTextures Annotation}},
        rounded corners, 
        width=\textwidth,
        enlarge left by=0mm,
        enlarge right by=0mm,
        boxrule=0.8pt
    ]
    
    \textbf{System Prompt:} \\
    You are an expert in face tampering detection. Your task is to strictly compare the \texttt{[Manipulated Image]} and the \texttt{[Original Image]}, first precisely locate the manipulated region (entire face), and then \textbf{explain} how the forgery artifacts were \textbf{caused by the NeuralTextures GAN-based full-face re-rendering process}. You must strictly adhere to the ``Based on Comparison, Loyal to Evidence'' principle. \textbf{Hallucinations are strictly forbidden}. Focus on artifacts left by the \textbf{forgery process} (e.g., loss of high-frequency details), not natural facial features. Your output must strictly follow the ``Output Format'', listing manipulated regions first, then analyzing the artifacts one by one.
    
    \vspace{2mm}
    \textbf{User Prompt:} \\
    \textbf{Task Definition:} You will receive two images: the first is a \texttt{[Manipulated Image]} generated using NeuralTextures technology, and the second is the corresponding \texttt{[Original Image]}. Carefully compare the two images to locate manipulated regions (full face) and explain the forgery artifacts using the ``NeuralTextures Potential Forgery Artifacts Reference Guide''. Focus solely on artifacts caused by NeuralTextures technology.
    
    \vspace{1mm}
    \textbf{Core Analysis Principles:}
    \begin{enumerate}
        \item Identify all differences across the full face, including skin texture, wrinkles, smile lines, and lips. Pay special attention to the lips and surrounding areas.
        \item Distinguish between:
        \begin{itemize}
            \item \textbf{Analyze This - Forgery Artifacts}: Pixel-level anomalies caused by GAN re-rendering.\\
            \textit{Correct Example}: ``Airbrushed'' feel, loss of high-frequency details (pores, wrinkles, smile lines), loss of lip texture, blurry lip borders.
            \item \textbf{Ignore This - Semantic Changes/Natural Features}: Expression changes are not artifacts.\\
            \textit{Incorrect Example}: Open mouth, smiling vs laughing.
        \end{itemize}
        \item Attribute observed phenomena to technical flaws according to the Reference Guide.
        \item Note: NeuralTextures preserves identity; focus on expression and full-face re-rendering artifacts.
    \end{enumerate}
    
    \vspace{1mm}
    \textbf{NeuralTextures Potential Forgery Artifacts Reference Guide:}
    \begin{enumerate}
        \item \textbf{Core Artifact: Global High-Frequency Detail Loss}
        \begin{itemize}
            \item \textit{Cause}: GAN/U-Net reconstruction of the entire face smooths out high-frequency details.
            \item \textit{Manifestation}: 
            \begin{itemize}
                \item ``Airbrushed'' feel across the entire face (forehead, cheeks, nose).
                \item Loss of skin texture, pores, smile lines, crow's feet; details are faded or missing.
            \end{itemize}
        \end{itemize}
        \item \textbf{Local Artifact: Mouth Detail Blurring}
        \begin{itemize}
            \item \textit{Cause}: GAN may struggle to preserve fine lip textures.
            \item \textit{Manifestation}: Loss of lip texture, blurry lip borders, stiff or flat lip shape.
        \end{itemize}
    \end{enumerate}
    
    \vspace{1mm}
    \textbf{Output Format:} \\
    First list manipulated regions, then describe corresponding forgery artifacts.\\
    \textit{Reference Example:} \\
    \textit{Manipulated Regions: Facial Skin, Lips\\
    Forgery Artifacts: [Facial Skin]: Entire face looks abnormally smooth, ``airbrushed'', lacks pores and texture. [Lips]: Lip surface abnormally smooth, missing lip texture.}

    \vspace{2mm}
    \textbf{Here are \texttt{[Manipulated Image]} and the \texttt{[Original Image]}.}\\
    \texttt{[Manipulated Image]}: \{\{manipulated\_image\_path\}\} \\
    \texttt{[Original Image]}: \{\{original\_image\_path\}\}

    \end{tcolorbox}

\end{figure*}

\begin{figure*}[t]
    \centering
    \begin{tcolorbox}[
        colback=gray!5,
        colframe=black,
        title={\textbf{Prompt for Real Image Annotation}}, 
        arc=2mm,
        width=\linewidth,
        boxrule=0.8pt,
        fontupper=\small, 
        top=2mm, bottom=2mm, left=2mm, right=2mm
    ]
    
    \textbf{System Prompt:} \\
    You are an expert in face tampering detection. Your task is to strictly analyze the \texttt{[Real Image]} and \textbf{``provide evidence''} of its authenticity. You must accomplish this task by \textbf{selectively} confirming that the image \textbf{possesses ``Indicators of Authenticity''} (e.g., normal facial structure) and \textbf{lacks ``Forgery Artifacts''} (e.g., no splicing lines on the face). Your analysis must be \textbf{based on facts}: if a real, low-resolution image is blurry, you \textbf{must not} label ``clear pores,'' but should instead focus on analyzing other indicators like ``lighting consistency'' or ``facial structure''. Your output must \textbf{strictly follow} the ``Output Format'' specified by the user.
    
    \vspace{2mm}
    \textbf{User Prompt:} \\
    \textbf{Task Definition:} \\
    You will receive a \texttt{[Real Image]}. Please act as an expert in face tampering detection to \textbf{analyze} and \textbf{explain} why this image is real. Your goal is to teach a downstream model to recognize \textbf{Indicators of Authenticity}. Hallucinations are strictly forbidden during this process. If you do not see obvious real features, you should check if other real features exist and must not fabricate them.
    
    \vspace{1mm}
    \textbf{Core Analysis Principles (Important):}
    \begin{enumerate}[leftmargin=*, noitemsep, topsep=2pt]
        \item \textbf{Selective Annotation (Key Principle):} The ``Indicators of Authenticity Reference Guide'' is a \textbf{``checklist,''} not a ``requirement''.
        \begin{itemize}[leftmargin=1em, noitemsep]
            \item You \textbf{must} select \textbf{only} those features from the guide that are \textbf{clearly and distinctly visible} in this \texttt{[Real Image]}.
            \item \textbf{No Hallucinations}: If a real, low-resolution photo causes ``skin texture'' to be blurry, you \textbf{must not} select \texttt{[Skin Texture]}. Skip it and analyze reliable indicators like \texttt{[Skin Tone Consistency]} or \texttt{[Facial Structure]}.
        \end{itemize}
        \item \textbf{Analysis Order (From Easy to Hard):}
        \begin{itemize}[leftmargin=1em, noitemsep]
            \item \textbf{First (Check Face Swap):} Ensure obvious artifacts (feature structure, contour edges, skin tone) are \textbf{absent}.
            \item \textbf{Second (Check Reenactment):} Ensure subtle artifacts (skin texture, lip texture/shape) are \textbf{absent}.
        \end{itemize}
        \item \textbf{Core Difference (Consistency):} Forgery = \textbf{Inconsistent}; Real = \textbf{Consistent}. Your analysis should reflect this consistency.
    \end{enumerate}
    
    \vspace{1mm}
    \textbf{Indicators of Authenticity Reference Guide:} \\
    \textit{(Select and analyze only clearly observable features)}
    \begin{enumerate}[leftmargin=*, noitemsep, topsep=2pt]
        \item \textbf{Structure \& Contour (vs. FaceSwap/DeepFakes flaws)}
        \begin{itemize}[leftmargin=1em, noitemsep]
            \item \texttt{Normal Facial Structure}: Features do not show misalignment, distortion, or overlap.
            \item \texttt{Natural Contour Transitions}: Smooth transitions to neck/hair; \textbf{no} splicing artifacts, hard edges, or unnatural blurring.
        \end{itemize}
        \item \textbf{Lighting \& Color (vs. FaceSwap/DeepFakes flaws)}
        \begin{itemize}[leftmargin=1em, noitemsep]
            \item \texttt{Consistent Skin Tone/Lighting}: Base skin tone and lighting direction are consistent across face/neck. No color patches.
            \item \texttt{Natural Shadows/Highlights}: Physically plausible shape and softness of shadows/highlights.
        \end{itemize}
        \item \textbf{Geometric Shape (vs. Face2Face/NT flaws)}
        \begin{itemize}[leftmargin=1em, noitemsep]
            \item \texttt{Natural Lip Shape}: Lips have curvature; \textbf{lacks} stiff, over-smoothed, or puppet-like feel.
            \item \texttt{Coordinated Facial Muscles}: Muscle movements (e.g., smile lines) follow anatomical logic.
        \end{itemize}
        \item \textbf{High-Frequency Details (Depends on image clarity)}
        \begin{itemize}[leftmargin=1em, noitemsep]
            \item \texttt{Consistent Skin Texture}: \textbf{(Clear Image)} Pores/wrinkles visible and uniform. \textbf{(Blurry Image)} Blur level is consistent with background.
            \item \texttt{Clear Lip Texture}: \textbf{(Clear Image)} Visible texture and natural sheen; clear borders.
            \item \texttt{Hair/Teeth Details}: \textbf{(Clear Image)} Sharp strands of hair and teeth texture.
        \end{itemize}
    \end{enumerate}

    \textbf{Output Format:}
    
    \textit{Reference Example 1: High-Quality Image} \\
    Indicators of Authenticity: \\
    \textit{[Skin Texture Details]:} The skin texture on the cheeks and forehead is clear and consistent with the neck texture; under-eye bags are visible. \\
    \textit{[Lip Texture]:} The lip texture is clear and natural, with a natural sheen. \\
    \textit{[Lighting Consistency]:} The lighting direction on the face and neck is consistent, with no local anomalies.
    
    \vspace{1mm}
    \textit{Reference Example 2: Low-Quality/Blurry Image} \\
    Indicators of Authenticity: \\
    \textit{[Facial Structure]:} The facial features conform to anatomical structure, with no misalignment or overlap. \\
    \textit{[Contour Transitions]:} Although the image is blurry, the transition from the facial contour to the background is still natural, with no splicing lines. \\
    \textit{[Lip Shape]:} The lip shape is natural and not stiff; the shadow under the lower lip is natural.

    \vspace{2mm}
    \textbf{Here are the \texttt{[Real Image]}.}\\
    \texttt{[Real Image]}: \{\{real\_image\_path\}\}
    
    \end{tcolorbox}
    \label{fig:real_prompt}
\end{figure*}


\begin{figure*}[t]
    \centering
    \begin{tcolorbox}[
        colback=gray!5,
        colframe=black,
        title={\textbf{Prompt for Cross-Validation Chain-of-Thought}}, 
        arc=2mm, 
        width=\linewidth, 
        boxrule=0.8pt,
        fontupper=\small, 
        top=2mm, bottom=2mm, left=2mm, right=2mm 
    ]
    
    \textbf{System Prompt:} \\
    \textbf{Task Definition:} \\
    You are a world-class face tampering detection analyst and an expert in logical reasoning. Your task is to ``role-play'' and generate a \textbf{Cross-Validation Chain-of-Thought} to create fine-tuning data. You will receive:
    \begin{enumerate}[leftmargin=*, noitemsep, topsep=2pt]
        \item \texttt{Query\_Image} (Primary Evidence)
        \item \texttt{RAG\_Context} (Secondary Evidence: 5 retrieved reference annotations + similarity scores, possibly noisy)
        \item \texttt{Ground\_Truth\_Label}
    \end{enumerate}
    
    \vspace{2mm}
    \textbf{Your reasoning must follow three steps:}
    
    \begin{description}[leftmargin=*, style=unboxed, noitemsep, topsep=2pt]
        \item[\textit{1. Preliminary Visual Analysis:}] 
        \begin{itemize}[leftmargin=1em, noitemsep]
            \item Analyze the \texttt{Query\_Image} in detail.
            \item Describe only the visual facts that indicate support for the \texttt{Ground\_Truth\_Label} (e.g., ``skin color on the face is uniform'', ``an unnatural blur at the mouth's edge'').
            \item Do not mention or assume the Ground Truth. Only report visual evidence.
        \end{itemize}
        
        \item[\textit{2. RAG Reference Information Analysis:}] 
        \begin{itemize}[leftmargin=1em, noitemsep]
            \item Objectively report the 5 pieces of retrieved information.
            \item Identify:
            \begin{itemize}[leftmargin=1em, noitemsep]
                \item \textbf{Supporting Evidence (consistent with Step 1)}: e.g., {("Real: clear skin texture", 0.70)}
                \item \textbf{Opposing Evidence (contradicts Step 1)}: e.g., {("Fake: mouth artifact", 0.85)}
            \end{itemize}
            \item Provide an objective analysis of the RAG context, noting both support and noise.
        \end{itemize}
        
        \item[\textit{3. Fusion, Reasoning, and Decision:}] 
        \begin{itemize}[leftmargin=1em, noitemsep]
            \item Cross-verify visual facts from Step 1 against RAG evidence:
            \begin{enumerate}[leftmargin=1em, noitemsep]
                \item \textbf{Verify Opposing Evidence:} If RAG's {Fake} evidence cannot be confirmed visually, treat it as retrieval noise and discard it.
                \item \textbf{Verify Supporting Evidence:} If RAG's {Real} evidence can be confirmed visually, accept it as valid.
            \end{enumerate}
            \item Conclude by integrating Step 1 and Step 2:
            \begin{itemize}[leftmargin=1em, noitemsep]
                \item Visual analysis supports the Ground Truth.
                \item Supporting RAG evidence reinforces the conclusion.
                \item Opposing RAG evidence is discarded as noise.
            \end{itemize}
        \end{itemize}
    \end{description}

    \vspace{2mm}
    \hrule 
    \vspace{2mm}

    \textbf{Output Format:}
    
    \texttt{<Preliminary Visual Analysis>}: \\
    \textit{[Analyze the {Query\_Image} in detail. Only describe visual facts that lead to the \texttt{Ground\_Truth\_Label}.]} \\
    \texttt{</Preliminary Visual Analysis>}

    \vspace{2mm}
    \texttt{<RAG Reference Information Analysis>}: \\
    \textit{["The retrieval report provides the following information:} \\
    \textit{\textbf{Supporting Evidence:} \textit{[Cite 1-2 highest-scoring pieces consistent with Step 1]}} \\
    \textit{\textbf{Opposing Evidence:} \textit{[Cite 1-2 highest-scoring pieces contradicting Step 1]}} \\
    \textit{textbf{Analysis:} \textit{Objectively evaluate the RAG context."]}} \\
    \texttt{</RAG Reference Information Analysis>}

    \vspace{2mm}
    \texttt{<Fusion, Reasoning, and Decision>}: \\
    \textit{[Step 1 visual facts + cross-verification with Step 2:} \\
    \textit{1. Verify Opposing Evidence $\rightarrow$ discard if inconsistent} \\
    \textit{2. Verify Supporting Evidence $\rightarrow$ confirm if consistent} \\
    \textit{3. Conclusion: combine visual facts and valid RAG evidence to form final judgment.]} \\
    \texttt{</Fusion, Reasoning, and Decision>}

    \vspace{2mm}
    \texttt{<Answer>} \texttt{Ground\_Truth\_Label} \texttt{</Answer>}

    \end{tcolorbox}
    \label{fig:cross_validation_prompt}
\end{figure*}
\begin{figure*}[t]
    \centering
    \begin{tcolorbox}[
        colback=gray!5,
        colframe=black,
        title={\textbf{Prompt for Evidence-Guided Correction Chain-of-Thought}}, 
        arc=2mm, 
        width=\linewidth, 
        boxrule=0.8pt,
        fontupper=\small, 
        top=2mm, bottom=2mm, left=2mm, right=2mm 
    ]
    
    \textbf{System Prompt:} \\
    You are a world-class face tampering detection analyst and an expert in logical reasoning. Your task is to ``role-play'' and generate a \textbf{Cross-Validation Chain-of-Thought} to create fine-tuning data. You will receive:
    \begin{enumerate}[leftmargin=*, noitemsep, topsep=2pt]
        \item \texttt{Query\_Image} (Primary Evidence)
        \item \texttt{RAG\_Context} (Secondary Evidence: 5 retrieved reference annotations + similarity scores)
        \item \texttt{Ground\_Truth\_Label}
    \end{enumerate}
    
    You must act as a perfect student model that initially misjudges the image but allows the RAG report to correct its intuition. Your reasoning must follow three steps:
    
    \begin{description}[leftmargin=*, style=unboxed, noitemsep, topsep=4pt]
        \item[\textbf{1. Visual Analysis (Admit Confusion):}] Analyze the \texttt{Query\_Image} in detail. Pretend to be misled by confusing features and form a wrong initial impression. Do not mention the \texttt{Ground\_Truth\_Label}.
        
        \item[\textbf{2. RAG Analysis (Highlight Golden Evidence):}] Inspect the \texttt{RAG\_Context}. Identify the ``Golden Evidence'' consistent with the \texttt{Ground\_Truth\_Label} (e.g., high-score $> 0.8$, dominant quantity). Only report RAG contents without making a conclusion.
        
        \item[\textbf{3. Fusion and Decision (The ``Aha!'' Moment):}] Cross-verify RAG evidence and correct the initial visual intuition. Steps:
        \begin{enumerate}[leftmargin=1.5em, noitemsep]
            \item \textit{Re-examination:} Re-inspect the \texttt{Query\_Image} prompted by Golden Evidence.
            \item \textit{Verification (Aha! Moment):} Describe subtle artifacts now visible due to RAG guidance.
            \item \textit{Conclusion:} Declare the final judgment after RAG-guided re-evaluation.
        \end{enumerate}
    \end{description}
    
    \vspace{2mm}
    \hrule 
    \vspace{2mm}
    
    \textbf{Output Format:}
    
    \texttt{<Preliminary Visual Analysis>}: \\
    \textit{Describe the confusing visual features that may lead to a wrong initial judgment. E.g., ``I (the student) observe that the overall lighting of the \texttt{Query\_Image} seems consistent, and no obvious artifacts are immediately visible to the naked eye. At first glance, it seems to be a [Wrong Label].''} \\
    \texttt{</Preliminary Visual Analysis>}
    
    \vspace{2mm}
    \texttt{<RAG Reference Information Analysis>} \\
    \textit{``However, the RAG report provides high-confidence `Golden Evidence' that contradicts my initial impression:} \\
    \textit{1. \textbf{Golden Evidence (Corrective):} \textit{[Cite the 1-2 highest-scoring pieces of evidence consistent with the GT, e.g., (``Fake: subtle mouth artifact'', 0.91)]}} \\
    \textit{2. \textbf{Majority Consensus:} \textit{[Point out quantity distribution, e.g., ``Among the 5 retrieved pieces, 4 consistently point to [GT Label]'']}} \\
    \textit{3. \textbf{Analysis:} \textit{[Objectively evaluate RAG, e.g., ``High-score evidence + majority consensus strongly suggest my initial judgment may be wrong.'']''}} \\
    \texttt{</RAG Reference Information Analysis>}:
    
    \vspace{2mm}
    \texttt{<Fusion, Reasoning, and Decision>}: \\
    \textit{Describe how you cross-verify RAG evidence and correct your initial visual intuition:} \\
    \textit{1. Re-examination: Re-inspect the specific areas indicated by RAG.} \\
    \textit{2. Verification (Aha! Moment): Confirm subtle artifacts missed initially.} \\
    \textit{3. Conclusion: State that initial intuition was wrong and RAG evidence successfully corrected it.} \\
    \texttt{</Fusion, Reasoning, and Decision>}
    
    \vspace{2mm}
    \texttt{<Answer>} \texttt{Ground\_Truth\_Label} \texttt{</Answer>}
    
    \end{tcolorbox}
    \label{fig:correction_prompt}
\end{figure*}
\begin{figure*}[t]
    \centering
    \begin{tcolorbox}[
        colback=gray!5,
        colframe=black,
        title={\textbf{Prompt for Resilient Rejection Chain-of-Thought}}, 
        arc=2mm,
        width=\linewidth, 
        boxrule=0.8pt,
        fontupper=\small,
        top=2mm, bottom=2mm, left=2mm, right=2mm
    ]
    
    \textbf{System Prompt:} \\
    You are a world-class face tampering detection analyst and an expert in critical thinking. Your task is to ``role-play'' and generate a \textbf{Resilient Rejection Chain-of-Thought}. You will receive:
    \begin{enumerate}[leftmargin=*, noitemsep, topsep=2pt]
        \item \texttt{Query\_Image} (Primary Evidence)
        \item \texttt{RAG\_Context} (Secondary Evidence: 5 retrieved reference annotations, which may be \textbf{misleading} or \textbf{noisy})
        \item \texttt{Ground\_Truth\_Label}
    \end{enumerate}
    
    You must act as a rigorous expert who \textbf{rejects misleading external evidence}. Even if the RAG report has high similarity scores, if it contradicts the visual facts of the \texttt{Query\_Image}, you must identify it as ``Noise'' and discard it. Your reasoning must follow three steps:
    
    \begin{description}[leftmargin=*, style=unboxed, noitemsep, topsep=4pt]
        \item[\textbf{1. Visual Analysis (Independent Judgment):}] Analyze the \texttt{Query\_Image} thoroughly based on your internal forensic knowledge. Form an initial hypothesis based strictly on visual cues (e.g., skin texture, lighting consistency).
        
        \item[\textbf{2. RAG Analysis (Critical Scrutiny):}] Inspect the \texttt{RAG\_Context}. Identify evidence that \textbf{conflicts} with your visual analysis or contains hallucinations (e.g., RAG claims ``blurry mouth'' but the image is sharp).
        
        \item[\textbf{3. Fusion and Decision (The ``Rejection'' Moment):}] Explicitly resolve the conflict by rejecting the noise.
        \begin{enumerate}[leftmargin=1.5em, noitemsep]
            \item \textit{Conflict Detection:} State clearly that RAG evidence contradicts visual facts.
            \item \textit{Falsification:} Explain \textit{why} the RAG evidence is invalid (e.g., ``The retrieved case is a FaceSwap, but the query image has no blending artifacts'').
            \item \textit{Final Verdict:} Discard the RAG evidence and stick to your internal visual judgment.
        \end{enumerate}
    \end{description}
    
    \vspace{2mm}
    \hrule
    \vspace{2mm}
    
    \textbf{Output Format:}
    
    \texttt{<Preliminary Visual Analysis>}: \\
    \textit{Describe your independent visual observation. E.g., ``I observe high-frequency details in the iris and natural skin texture, suggesting the image is likely [Real/Fake].''}\\
    \texttt{</Preliminary Visual Analysis>}
    
    \vspace{2mm}
    \texttt{<RAG Reference Information Analysis>}: \\
    \textit{``The RAG report provides the following evidence, but it appears suspicious:''} \\
    \textit{1. \textbf{Conflicting Evidence:} \textit{[Cite high-scoring evidence that is factually wrong for this image]}} \\
    \textit{2. \textbf{Analysis:} \textit{[Note the discrepancy, e.g., ``Although the retriever gives a 0.85 score, the described artifacts are NOT present in the query image.'']}}\\
    \texttt{</RAG Reference Information Analysis>}
    
    \vspace{2mm}
    \texttt{<Fusion, Reasoning, and Decision>}: \\
    \textit{Perform the rejection reasoning:} \\
    \textit{1. \textbf{Conflict:} ``RAG suggests [Wrong Label] based on [Artifact], but my visual analysis shows [Visual Fact].''} \\
    \textit{2. \textbf{Rejection:} ``The retrieval is likely due to semantic similarity (e.g., similar pose) rather than forensic similarity. The evidence is a False Positive.''} \\
    \textit{3. \textbf{Conclusion:} ``I reject the misleading RAG evidence and trust my internal visual analysis.''}\\
    \texttt{</Fusion, Reasoning, and Decision>}
    
    \vspace{2mm}
    \texttt{<Answer>} \texttt{Ground\_Truth\_Label} \texttt{</Answer>}
    
    \end{tcolorbox}
    \label{fig:rejection_prompt}
\end{figure*}

\end{document}